\documentclass[10pt,twocolumn,letterpaper]{article}

\usepackage{cvpr}

\usepackage{multirow}
\usepackage{graphicx}
\usepackage{appendix}
\usepackage{wrapfig}
\usepackage{multicol}
\usepackage{algorithm,algorithmic}
\usepackage{colortbl}
\usepackage{xcolor}
\usepackage{array}
\usepackage{mathtools}

\newcommand{\mb}[1]{\mbox{$#1$}}
\def\m#1{\mathbf #1}
\DeclareMathOperator{\E}{\mathbb{E}}
\newcommand{\mcO}{\mathcal{O}}
\newcommand{\mcI}{\mathcal{I}}
\newcommand{\mcD}{\mathcal{D}}
\newcommand{\mcP}{\mathcal{P}}
\newcommand{\xhe}{\ensuremath{x'_\textsc{h}}}
\newcommand{\xh}{\ensuremath{x_\textsc{h}}}
\newcommand{\xl}{\ensuremath{x_\textsc{l}}}
\newcommand{\The}{\ensuremath{t'_\textsc{h}}}
\newcommand{\rhe}{\ensuremath{r'_\textsc{h}}}

\newcommand{\ttsize}{%
    \fontsize{7}{11}\selectfont 
}
\def\tt#1{\texttt{\ttsize #1}}
\def\ttbig#1{\texttt{\small #1}}

\makeatletter
\renewcommand{\ALG@name}{Function}
\makeatother

\def\fref#1{Func.\hspace{0.2em}\ref{#1}}

\definecolor{cvprblue}{rgb}{0.21,0.49,0.74}
\usepackage[pagebackref,breaklinks,colorlinks,allcolors=cvprblue]{hyperref}

\def\paperID{1501}
\def\confName{CVPR}
\def\confYear{2025}

\title{Removing Reflections from RAW Photos}

\author{
\begin{tabular}{c}
Eric Kee \and
Adam Pikielny \and
Kevin Blackburn-Matzen \and
Marc Levoy
\end{tabular} \\
\centering 
\begin{tabular}{c}
Adobe Inc. \\
{\tt\small \{kee,pikielny,matzen,levoy\}@adobe.com}
\end{tabular}
}

\begin{document}
\twocolumn[{%
\renewcommand\twocolumn[1][]{#1}%
\maketitle
\vspace{-1.5em}
\begin{center}
    \centering
    \captionsetup{type=figure}
    \includegraphics[width=0.7\linewidth]{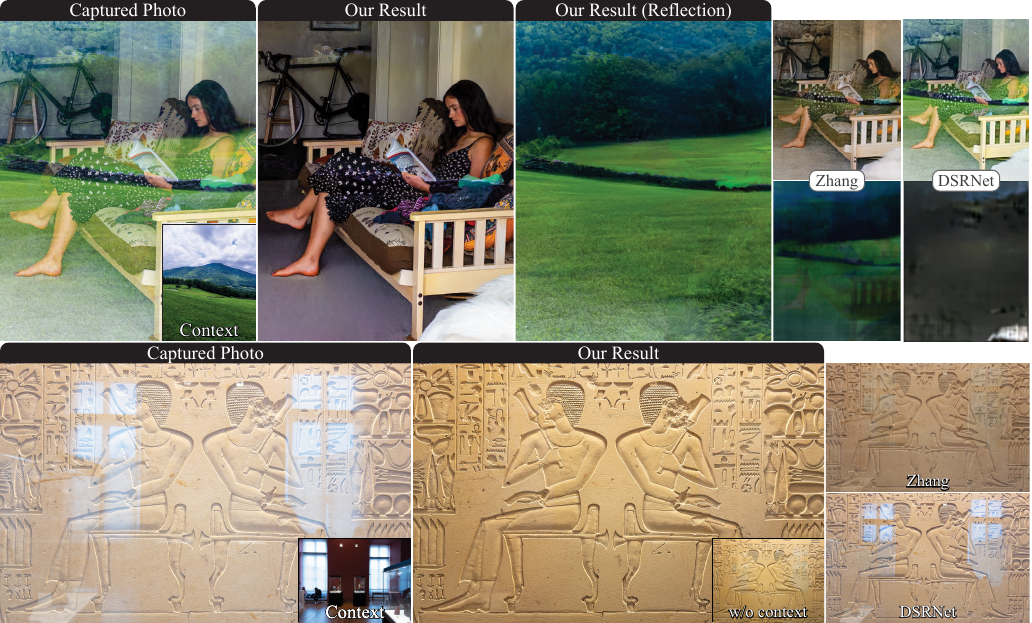}
    \captionof{figure}{Results of our reflection removal system. We use linear (RAW) images with an optional contextual photo, and output the clean and reflection images in linear color for editing, at full resolution (shown at $2\mathrm{K}$). Prior works use tone-mapped images at $\approx 256$p, yielding lower quality and inaccurate color. Brightness/contrast changes relative to captured photos arise from reflection removal, and are correct.}
    \label{fig:main_result}
\end{center}%
}]

\maketitle

\begin{abstract}
We describe a system to remove real-world reflections from images for consumer photography. Our system operates on linear (RAW) photos, and accepts an optional contextual photo looking in the opposite direction (\eg the ``selfie'' camera on a mobile device). This optional photo disambiguates what should be considered the reflection. The system is trained solely on synthetic mixtures of real RAW photos, which we combine using a reflection simulation that is photometrically and geometrically accurate. Our system comprises a base model that accepts the captured photo and optional context photo as input, and runs at 256p, followed by an up-sampling model that transforms 256p images to full resolution. The system produces preview images at 1K in 4.5-6.5s on a MacBook or iPhone 14 Pro. We show SOTA results on RAW photos that were captured in the field to embody typical consumer photos, and show that training on RAW simulation data improves performance more than the architectural variations among prior works.%
\end{abstract}%
\vspace{-1.5em}
\section{Introduction}\label{sec:introduction}

Taking pictures through glass is difficult. Light reflects off of the glass and linearly mixes with the subject, creating a distraction. Photos from cars and airplanes show the cabin, photos from buildings include the ceiling lights, paintings are covered by haze, and window shopping shots are photo-bombed by the photographer---to name just a few cases.

Removing unwanted reflections is difficult because they occur in a diverse range of locations and situations. Locations include {\em shopping} spots, {\em traveling} (cars, planes), buildings, museums, and special cases (eyeglasses, screens). Reflections depend on the time, lighting (\eg incandescent), scene (trees, streets), illuminant power, and appearance (complex textures or simple shapes). These factors create priors because glass is placed carefully in the world.

One way to remove reflections involves capturing a second photo with a black material placed behind the glass to allow only reflected light to reach the camera. If this {\em reflection image}, and the original {\em mixture image}, are stored in a format that preserves the linear relationship between pixel values and the scene luminance (e.g., RAW), then these two {\em scene-referred} images can be subtracted to obtain the image that transmitted through the glass. This {\em transmission image} can be recovered because light mixes by addition in photosites on the sensor. Subtraction has been used for datasets~\cite{lei2021b, wan2022}, but fails under motion or lighting changes; this severely restricts data collection. Alternatively one can place glass panes in the scene, but the scene and lighting are typically similar on both sides. Training and evaluating on such unrealistic data can be unhelpful and misleading.

This paper presents a reflection removal system for consumer photography that targets the following requirements:
\begin{enumerate}
	\item Handle typical reflections in consumer photography.\label{req:typical_reflections}
	\item Minimize user interactions (steps, taps, strokes).\label{req:typical_steps}
	\item Allow photo capture in a typical amount of time.\label{req:typical_time}
	\item Produce results on-screen for review in about 5 seconds.\label{req:fast_review}
	\item Produce results at the input image resolution.\label{req:full_resolution}
	\item Facilitate editing for error correction and aesthetics.\label{req:error_correction}
\end{enumerate}

Few prior works satisfy these, which affect design and evaluation. In particular, data should match how the system will be used. 
For req.~\ref{req:typical_reflections}, one needs a large dataset of realistic photos. Prior works require ground truth photos, but capturing them restricts the dataset size and diversity. We synthesize realistic and diverse photos in large quantities.

We synthesize reflections by combining, for example, an image looking at a storefront with one of sunlit buildings presumed to be behind the photographer. To make the synthesis accurate, we use linear scene-referred images with known photometric and colorimetric calibration, and combine them in a physically correct way. For example, the reflected buildings are typically brighter and bluer than the storefront, but will be attenuated by reflection off the glass.

To address req.\ \ref{req:typical_steps} (capture time) and \ref{req:fast_review} (processing time), we avoid asking the user to capture video, bursts of frames, or stereo photos.  These help identify what created the reflection, but they slow down processing. Instead, we allow the user to capture an optional, contextual photo. This photo does not need to be captured simultaneously, or registered with the original photo.  In fact, it could be captured by quickly turning around and looking away from the window.

To address req.\ \ref{req:full_resolution}, we use a novel upsampler with a flexible output resolution. Note that upsampling is imperative and non-trivial, but mostly disregarded in the literature. To meet req.\ \ref{req:error_correction} we output reflection and transmission images so users can remix them to fix the long tail of practical failures.

\noindent\textbf{Contributions.} In this work, we 
\begin{enumerate}
    \item show how to synthesize training data such that models do not need to be fine-tuned on real images;
    \item show that training/testing on RAW improves performance significantly---more than prior model variations;
    \item use a contextual photo to help identify the reflection;
    \item significantly reduce upsampling artifacts while producing output at 1K in 5s for review, and at full resolution.
\end{enumerate}
{\bf This paper is best read with hyperlinks into the supplement. See arXiv or the project website for a complete version.} Prior work is outlined in \cref{sec:prior_work}, reflection synthesis (\cref{sec:reflection_synthesis}), removal (\cref{sec:reflection_removal}), and results (\cref{sec:results}). In supplemental sections we discuss simulation (\cref{sec:supp:photometric_sim}--\ref{sec:supp:contextual_photo}), data collection (\cref{sec:supp:data_collection}), modeling (\cref{sec:supp:reflection_removal}), and results (\cref{sec:supp:results}). 

\section{Prior work}\label{sec:prior_work}

Removing reflections is a long-standing problem.  Prior works have used multi-image capture and machine learning. Among the latter, upsampling low-resolution results is an important sub-problem. We survey each category.

\textbf{Multiple input images.} Prior methods use video~\cite{alayrac2018, guo2014}, image sequences~\cite{han2017, li2013, liu2020, sarel2004, sarel2005, sun2016, szeliski2000, xue2015}, flash~\cite{agrawal2005, lei2021}, near infrared~\cite{hong2020}, polarization~\cite{farid1999, kong2014, lyu2019, wieschollek2017, lei2020}, and dual pixel images~\cite{punnappurath2019}, as well as light fields~\cite{wang2015}.
We use an optional and additional photo of the reflected scene (not of the glass) to identify the reflection. This {\em contextual photo} is any for which the camera is pointed at the reflected scene (e.g., the camera is turned $180^\circ$ as in a ``selfie'' camera).

\textbf{Reflection synthesis.} 
Prior methods are trained with heuristically mixed pairs of tone-mapped images~\cite{chen2024, qiming2023, guo2014, li2014, yang2018, wan2020, yang2019, jin2018, arvanitopoulos2017, fan2017}.
Such mixing is inaccurate, so non-linear methods have been used~\cite{wen2019, kim2019}. Physically based methods nonetheless use tone-mapped images~\cite{kim2019}. Successful methods however require ground truth images to train models that generalize~\cite{lei2021b}, typically at approximately a $10$:$1$ ratio of synthetic and real~\cite{qiming2023, lei2021b, wan2020, zhang2018, wei2019, li2019, prasad2021, li2020}. This ratio raises issues of dataset scale and diversity because ground truth capture is tedious and restrictive. The largest dataset of real images to-date~\cite{zhu2023} has 14,952 pairs ($10^4$), but methods like~\cite{qiming2023, wan2020, zhang2018} require pre-training on datasets larger than $10^6$ (e.g., ImageNet~\cite{simonyan2015}). 
We synthesize photometrically accurate images to obviate ground truth training images, and train models from scratch on more than 1M examples, which improves performance. 

\textbf{Removing high resolution reflections.} Most methods operate at $\approx\hspace{-0.3em}256^2$ pixels, and cannot be trivially scaled up. Useful systems must create preview images at $\approx\hspace{-0.3em}1\mathrm{K}$ pixels, and final outputs beyond $4\mathrm{K}$. Prasad~\cite{prasad2021} use a base model at $256^2$ pixels, and an upsampler that yields $\ge\hspace{-0.3em}4\mathrm{K}$ pixels. Their fast upsampler re-introduces sharp reflections. 
Our upsampler is similarly fast, but removes sharp reflections.

\textbf{Inference on RAW images.} Most prior methods apply reflection removal to $8$-bit {\em display-referred} images, such as internet JPEGs. 
Such images have been white-balanced, tone-mapped, denoised, sharpened, and compressed.  We reframe dereflection to operate on scene-referred (RAW) images. Lei~\cite{lei2021} subtract pairs of RAW images to suppress the reflection before converting to $8$-bit for full removal. We operate on RAW end-to-end. RAW inputs improve prior methods, but our system outperforms them.

\section{Reflection synthesis}
\label{sec:reflection_synthesis}

Our pipeline for removing reflections uses a base model and an upsampler that are trained solely on simulated images (\cref{sec:reflection_removal}), which overcomes the scaling bottleneck of needing to capture real reflections. We simulate reflections photometrically by summing pairs of {\em scene-referred} images, which are linear with respect to scene luminance. In contrast, images in most $8$-bit formats are {\em display-referred}---non-linearly related to luminance. Scene-referred images originate from sensor data stored in RAW format, such as Adobe Digital Negative (DNG). The transformation of RAW data into display-referred images is described by Adobe Camera RAW (ACR), the DNG spec.~\cite{acr} pp.99-104, and the DNG SDK~\cite{acr} as follows:
\begin{enumerate}
    \item Linearize (e.g. remove vignetting and black levels) \label{acr:linearize}
    \item Demosaic \label{acr:demosaic}
    \item Subtract the scalar black level \label{acr:subtract_black}
    \item Convert to XYZ color \label{acr:convert_XYZ}
    \item White balance\label{acr:white_balance}\footnote{ACR defines two white balancing paths, and we leverage one that differs from many cameras and the literature~\cite{afifi2022, brooks2019, karaimer2016}. In the literature, white balance is applied before converting to XYZ with the {\em forward matrix}. ACR also supports that ordering (DNG Spec.\cite{acr} p103, matrix \texttt{FM}), but reflection simulation requires the opposite (as explained in \cref{sec:reflection_synthesis}). Fortunately, ACR specifies a second path that uses {\em color matrices} (DNG Spec.~\cite{acr} pp101-103, matrix \texttt{CM}), to transform to XYZ before white balancing.  All DNGs are required to provide such color matrices, whereas the forward matrices of the first path are optional. ACR recommends forward matrices under extreme lighting (DNG Spec.~\cite{acr} pp.101-103), for which they are more precise. Both paths however depend on the as-shot illuminant; see ACR Funcs.~\ref{func:calc_whitexy}, \ref{func:cam_to_rgb}, \ref{func:find_xyz_to_cam}. 
In \cref{sec:results}, we show that this color processing yields synthetic training data with sufficient realism for models to generalize to photos in-the-wild from other cameras, while prior methods do not.
}
    \item Convert to RGB color \label{acr:convert_rgb}
    \item Dehaze, tone map (spatial adaptive highlights, shadows, clarity);
enhance texture; adjust local contrast, hue, color tone, whites, and blacks.
    \label{acr:tone_mapping}
    \item Gamma compress \label{acr:gamma_compression}
\end{enumerate}
Step~\ref{acr:gamma_compression} yields an $8$-bit {\em finished image} for storage, but its pixel values are non-linearly related to scene luminance because Step~\ref{acr:tone_mapping} performs proprietary, non-linear, and spatially varying effects that cannot be modeled with a gamma curve as is often done~\cite{zhang2018, wieschollek2017, li2019}.  {\em Realistic reflections therefore cannot be simulated by summing pairs of finished images.} 

Which earlier step is most appropriate for simulation? The outputs of Steps~\ref{acr:white_balance} and \ref{acr:convert_rgb} are linear, but the illuminant color has been removed by white balancing---accurate reflections cannot be simulated here because scenes that reflect from and transmit through glass are often illuminated by light sources with differing colors, and those colors mix before white balancing. The output of Step~\ref{acr:subtract_black} is linear, preserves the illuminant color, and has been demosaicked, but its colors are with respect to a sensor-specific spectral basis---images from different sensors cannot be summed here. The output of Step~\ref{acr:convert_XYZ} is however ideal: the XYZ color space is sensor-independent, the illuminant color is preserved (unlike prior works \cite{afifi2022}), and pixels are linear with respect to luminance. We therefore select Step~\ref{acr:convert_XYZ} and XYZ color space to simulate photometrically accurate reflections. 

\subsection{Photometric reflection synthesis}\label{sec:photometric_synthesis}

Our most fundamental simulation principle is the additive property of light: glass superimposes the light fields from a reflection and transmission scene to form a mixture. 
The resulting mixture image $m = t + r$ accumulates (with equal weight) photons from the two scenes into a transmission image $t$ and a reflection image $r$. We simulate $t$ and $r$ from images in linear XYZ color (ACR Step~\ref{acr:convert_XYZ}). 

The first photometric property is the illuminant color, which often differs between $t$ and $r$ because the glass in consumer photographs typically separates indoor and outdoor spaces. Otherwise, the photographer could walk around the glass to take their photo. Even in specialized scenes like museum display cases, the case is often internally illuminated, making its illuminant color different than in the gallery at large. By representing $t$ and $r$ in XYZ color before white balancing, the illuminant colors are mixed.

The second property is the illuminant power. In typical scenes, this power differs on either side of the glass ($t$ and $r$ differ in brightness). The number of captured photons is scaled by the exposure $e = s\cdot g / n^2$, for shutter speed $s$, aperture $n$, and gain $g$ (ISO). We normalize the exposures of $t$ and $r$ by $e$ so pixels are proportional to scene luminance up to a shared constant. This un-exposed mixture is $m' = t' + r'$, $t' = t/e_t$, and $r' = r/e_r$, for exposures $e_t$ and $e_r$. We simulate a capture function $\mathcal{C}$ that re-exposes and re-white balances $m'$ by exposing the mean pixel to a target value $\tau$, $m = \mathcal{C}(m') = We'm'$, and $e' = \tau / \E[m']$, where $W$ is a $3\times3$ matrix that white balances in XYZ (\fref{func:white_balance}, \cref{sec:supp:white_balancing}). If pixels in $t$ or $r$ are saturated, \mb{e' = 1 / \min(\max(t'),~\max(r'))}, to ensure they remain so. Lastly, $m$ is converted to scene-referred, linear RGB to train models. 

The full simulation is described in \fref{func:simulate_examples} and \cref{sec:supp:photometric_sim}. This function produces mixtures $m$ are photometrically accurate, but they aren't always useful. When saturation dictates the re-exposure $e'$, pixels can be clipped, modeling over-exposed $m$. Images $t'$ or $r'$ can also be so dark that they are invisible, or so mutually destructive that one would struggle to identify the subject.
These photos do not model $m$ that photographers care about. We therefore collect a large dataset of images and search for well exposed and well mixed $m$. This search introduces photometric and semantic priors on $m$, $t$, and $r$ (e.g., skies often reflect). See \cref{sec:supp:data_collection}.

\subsection{Geometric reflection synthesis}\label{sec:geometric_sim}

Our second fundamental simulation principle is that mixtures must be geometrically valid. Denoting the images to be summed as $t$ and $r$, and our source image pairs as $(i,j)$, we synthesize \mbox{$t = T(i)$} and \mbox{$r = R(j)$} by modeling spatially varying Fresnel attenuation, perspective, double reflection, and defocus. We omit from $T$ the effects of global color, dirt, and scratches; editing tools can correct them. We model a physically calibrated amount of defocus blur; most reflections are sharp as also noted in~\cite{lei2021b}. See \cref{sec:supp:geometric_sim}.

\subsection{The contextual photo}\label{sec:contextual_photo}

We accept an optional contextual photo $c$ that directly captures the reflection scene to help identify the reflection $r$. Capture of $c$ can be simultaneous with the secondary {\em front camera} (selfie) on a mobile device, or briefly later. We make three observations about the views of $c$ and $r$ (see \cref{fig:supp:selfie_geometry}):
\begin{enumerate}
	\item Even if the cameras are collocated, the viewpoints of $c$ and $r$ will be translated by twice the distance to the glass.
	\item If the mixture is captured obliquely to the glass, rotating the contextual view 180$^\circ$ yields little common content.
	\item If the selfie camera is used, the reflection scene might be partially occluded by the photographer. 
\end{enumerate}
Image $c$ will therefore often contain little content that matches with $r$ unless it is captured carefully. We avoid placing such a large burden on the user, and allow them to capture any view, $c$, of the reflection scene. Crucially, this relaxation also facilitates the geometric simulation. We scalably model $c$ by cropping source images into a disjoint left/right half (or top/bottom). 
The context image encodes information about the lighting and scene because we use a capture function $\mathcal{C}$ with the same white balance as $(m, t, r)$. See~\cref{sec:photometric_synthesis}, \fref{func:simulate_examples}, and \cref{sec:supp:contextual_photo} for details.

\begin{algorithm}[t]
\caption{Simulate reflection examples $(m, t, r, c)$.}
\footnotesize
\textbf{Input:} A random pair of XYZ images $(i, j)$ \\%
\hspace{0em}\mbox{\textbf{Output:} Simulated components and context image.}
\vspace{-1.25em}
\begin{algorithmic}[1]
    \STATE Split $j$ into non-overlapping reflection and context parts $(r, c)$.%
    \STATE Split $i$ similarly: randomly select a transmission part $t$.
    \STATE Unexpose $t$ and $r$ by using their exposure metadata.
    \STATE Apply the geometric simulation to $(t, r)$.
    \STATE Composite $m = t + r$.
    \STATE Compute a new exposure $e$ for $m$. \hfill\COMMENT{\fref{func:compute_exposure}}
    \STATE Compute WB matrix \tt{XYZ\_to\_XYZ\_awb}. \hfill\COMMENT{\fref{func:white_balance}}
    \STATE White balance (WB) $m$ by applying \tt{XYZ\_to\_XYZ\_awb}.\label{func:simulate_examples:calc_wb}
    \STATE Apply the same white balance to $(t, r, c)$.\label{func:simulate_examples:wb_trc}%
	\STATE Get the transform \tt{XYZ\_D50\_to\_sRGB}. \hfill\COMMENT{SDK \fref{func:xyz_d50_to_srgb}}
	\STATE Transform $(m, t, r, c)$ to linear sRGB.
	\RETURN $(m, t, r, c)$
\end{algorithmic}
\label{func:simulate_examples}
\vspace{-0.25em}
\end{algorithm}

\section{Reflection removal}\label{sec:reflection_removal}

Our system removes reflections from RAW images, $m$, in linear RGB color (ACR Step~\ref{acr:convert_rgb}) with an optional context image $c$ that is white balanced like $m$ (see \fref{func:simulate_examples}). Both $m$ and $c$ share a scene-referred color space, which aids removal; RGB supports pre-trained perceptual losses. We predict $t$ and $r$ in linear RGB, and store inference outputs by inverting ACR steps \ref{acr:subtract_black}--\ref{acr:convert_rgb} to produce new RAW images.

\setlength{\columnsep}{1.5em}
\begin{wrapfigure}[11]{r}{0.3\linewidth}
  \centering
   \includegraphics[width=\linewidth]{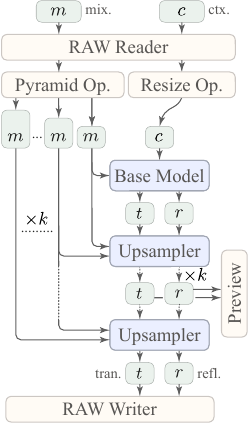}
   \vspace{-1.75em}
   \caption{System}
   \label{fig:system}
   \vspace{1.75em}
\end{wrapfigure}%
Our system uses two models, \cref{fig:system}. A base model uses $m, c$ at $256^2$ pixels to predict $t$, $r$ (rectangular images are tiled); $t, r$ are then upsampled using a Gaussian pyramid. 

\subsection{Base model}\label{sec:base_model}

The base model is in \cref{fig:base_model} due to space limits. A multi-scale backbone projects $m$ into a high dimensional space and computes semantic features (labeled {\em P-Net}). These features are fused (labeled {\em F-Net}) with a feature pyramid network (FPN) at the input resolution. The backbone is EfficientNet~\cite{effnet} at $256^2$ pixels, and fusion uses a BiFPN pyramid~\cite{effdet, timm}.

The context image $c$, is processed identically to $m$. Its low-resolution FPN features are used to predict affines that modify the FPN features of $m$ using conv-mod-deconv operations ala StyleGAN~\cite{karras2020}. Modulation is per-channel because $c$ does not share identical content with $m$. Conceptually, modulation gives the model additional capacity to identify $r$ within the features of $m$. A finishing module further identifies and renders $t, r$ (it's the head in~\cite{zhang2018}). We predict $(t, r)$ independently, rather than enforcing $t + r = m$, to decouple failures. Training uses the losses of~\cite{zhang2018} with improvements to the adversary and gradient terms. Crucially, training is end-to-end from random weights. See \cref{sec:supp:base_model}.%

\subsection{Upsampler}\label{sec:upsampler}

The upsampler is shown in \cref{fig:upsampler} due to space limits. Upsampling is performed iteratively over a Gaussian pyramid (see \cref{fig:system}), as summarized below. Details are in \cref{sec:supp:upsampler}. 

Briefly, the upsampler first projects the low- and high-resolution images $(m, r, t)$, and $M$ into a high dimensional space $\phi$ using a convolutional backbone. The upsampler then matches low resolution features $\phi_t$, $\phi_r$ to $\phi_m$ to create masks that identify the features of $t$, $r$ within $m$. This matching process uses products of features: when features match, their product can be large regardless of sign, whereas summation yields large activations if either input is large. We generalize this idea by predicting affine transforms that are applied to the features of $t$ and $r$, followed by a sigmoid; see \cref{fig:upsampler} (bottom). Two per-pixel, per-channel masks are thus predicted, $\mathbb{I}_t$, $\mathbb{I}_r$. Errors are corrected by a joint mask predictor that inspects both $\mathbb{I}_t$, $\mathbb{I}_r$ (see \cref{sec:supp:upsampler}). Masks $\mathbb{I}_t$ and $\mathbb{I}_r$ are resampled $2\times$ and multiplied with $\phi_M$ to project its features into subspaces for $T$, $R$. This key step assumes that the identity $\mathbb{I}_t$, $\mathbb{I}_r$ of the component to which each feature belongs is low in spatial frequency. By resampling masks, not features, sharp features are preserved. Errors are corrected with finishing convolutions, which render $T,R$.

Training uses a cycle-consistency loss, losses similar to~\cite{prasad2021}, and begins from scratch. See \cref{sec:supp:upsampler} for details.

\section{Results}
\label{sec:results}

We evaluate the simulation, base model, and upsampler; extensive results are added in the supplement. We make four contributions. {\em First}, we show that dereflection models that are trained solely on simulated reflections can generalize to real images without fine-tuning on real images, provided that the simulation uses RAW images in a photometrically accurate way. {\em Second}, training and testing on RAW images improves results significantly, and more than prior model variations. {\em Third}, a context image can disambiguate the reflection content if it is captured in another direction (e.g., the selfie). {\em Fourth}, upsampling low-res results, which is imperative but largely neglected in the literature, works better if one explicitly matches features in the low-res outputs $t, r$ to the low-res mixture, and masks them from the high-res mixture to recover the high-res transmission and reflection.

\subsection{Reflection simulation}\label{sec:results_dataset}

\begin{figure*}[t!]
    \begin{minipage}[t]{0.651\linewidth}    
        \centering
        \vspace{0pt}
	   \includegraphics[width=\linewidth]{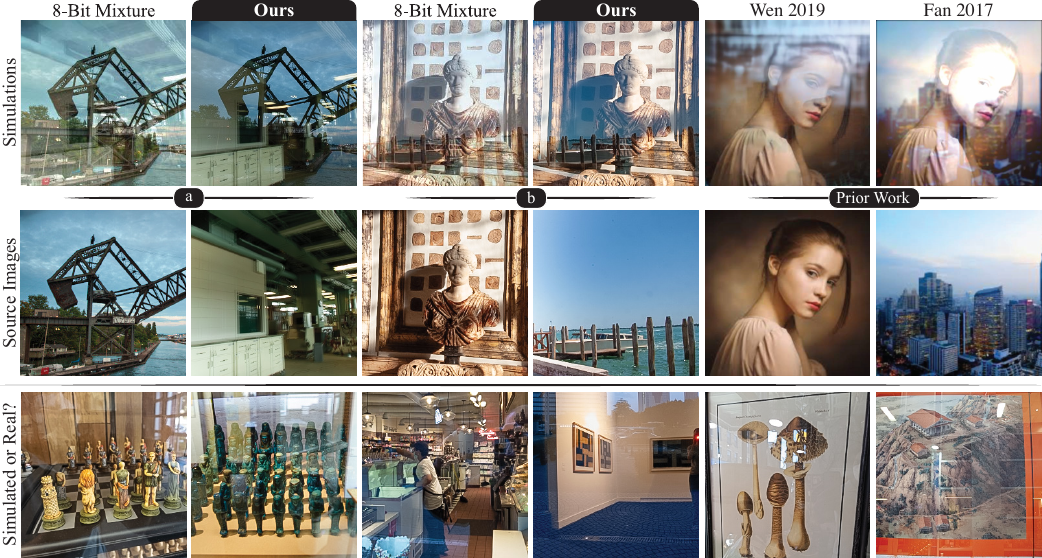}
	   \caption{The importance of synthesizing training data (top row) from linear images (middle row), compared to prior work. {\bf (a)} Photometrically accurate illuminant colors are simulated by mixing before white balancing; mixing 8-bit white balanced images is much different. {\bf (b)} Mixing in scene-referred linear units produces reflections that are strong in the shadows, but transparent in the highlights. {\bf (prior work)} Such effects are visibly incorrect in prior work, which blend $8$-bit tone mapped images~\cite{wen2019, fan2017}. {\bf (bottom)} Real and simulated examples are shuffled together. For each real image, a similar synthetic reflection was manually found in the dataset. Real images were not captured to match known examples; these qualitative matches exist because the dataset size exceeds $10^6$ 
		\raisebox{0.45em}{{\rotatebox{180}{(even numbered images are synthetic)}}}.
		}
       \label{fig:dataset_photometric}
    \end{minipage}
    \hfill
    \begin{minipage}[t]{0.334\linewidth}    
        \centering
        \vspace{0pt}
	    \includegraphics[width=\linewidth]{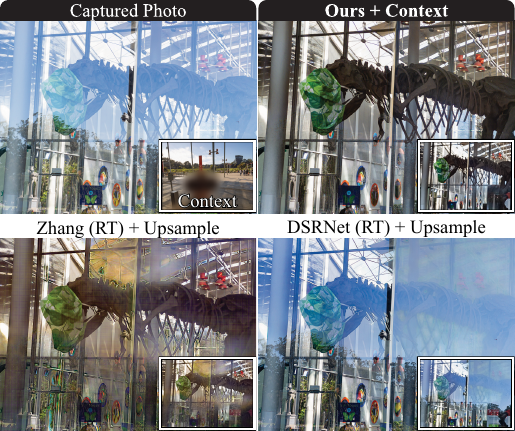}   
        \vspace{-1.75em}
        \caption{Results at $2048$p; base outputs inset.}
	    \label{fig:extra_result}
        \vspace{0.25em}        
	    \includegraphics[width=\linewidth]{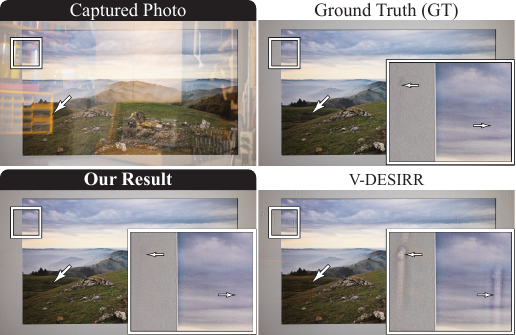}   
        \vspace{-1.75em}
        \caption{Upsampling GT images $256$p to $2048$p. V-DESIRR~\cite{prasad2021} adds artifacts.}
        \vspace{0.25em}        
	    \label{fig:results_upcompare}
    \end{minipage}
\end{figure*}

Source images were drawn from MIT5K~\cite{mit5k}, RAISE~\cite{raise}, and Laval Indoors~\cite{gardner2017}, totaling $12$,$803$ RAWs and $2$,$233$ scene-referred Image-Based Lighting (IBL) panoramas. The $360^\circ$ IBLs are equivalent to about $12$,$367$ indoor RAW images because we simulate random cameras with an average FOV of $65^\circ$, \cref{sec:supp:camera_projection}, \ref{sec:supp:hdr_env_sampling}. Images are grouped into $10$,$547$ outdoor and $14$,$623$ indoor to create pairs \mb{(i,j)}, \cref{sec:supp:image_collection}. The groups are split into train, validation, and test sets (80\%, 15\%, 5\%) before simulation.

The number of examples $(m,t,r)$ is amplified by randomization in the geometric simulation (\cref{sec:supp:geometric_sim}, \ref{sec:supp:data_collection}). We search $10^8$ examples for useful $m$. After culling, about $10^7$ mixtures remain, and we rendered $10\%$ at $256^2$ and $2048^2$ pixels to train the models. The $256^2$ pixel dataset has $1$,$241$,$091$ for training, $46$,$121$ for test, and $8$,$991$ for validation; the $2048^2$ dataset has \mbox{$1$,$079$,$631$; $39$,$916$; and $7$,$448$.} 

\cref{fig:dataset_photometric} shows results of mixing scene-referred images: (a) correct illuminant colors and (b) correct reflection visibility. We linearly blended $8$-bit tone-mapped images for comparison, and compare to prior works (see caption). \cref{fig:supp:dataset_overview} shows an overview of the dataset, and is discussed below.

In \cref{tab:sim_ablation} we ablate each simulation component. We (\emph{gamma}) compressed $t$, $r$ before compositing; separately exposed them (\emph{exposure}); did not constrain their inclination or field-of-view (\emph{pose}); removed spatially varying attenuation by making all camera rays normal to the glass (\emph{fresnel}); separately white balanced (\emph{WB}); removed depth-of-field blur and double reflection (\emph{blur}); and removed all. We trained our base model on each dataset, and evaluated on the full-simulation test set. Each feature affects performance (all differences are significant), and omitting them all (\emph{All $\pm$ G}) decreases performance dramatically compared to the average degradation to $t, r$ (\emph{control}). 

\textbf{Discussion.} Prior works mix $8$-bit tone-mapped images, and the results are qualitatively unrealistic. Their simulated reflections overpower the highlights, and are not powerful enough in the shadows, which are boosted by tone mapping. In our accurate simulations, light from two scenes is mixed linearly and equally without tone-mapping. {\em This accurate mixing allows our models to generalize better to new scenes, and yields SOTA performance without training on real images (\cref{sec:base_reflection_removal}).}  Furthermore, by synthesizing physically accurate reflections {\em and searching for visible ones}, we introduce natural priors on their appearance. Indoor light is weak, so reflections of the indoors are typically of regions near light sources or windows; see \cref{fig:supp:dataset_overview}, examples $1$, $2$, $5$, $11$, $14$, $19$, $24$, $25$.  Indoor lights create small reflections that often look yellow atop outdoor scenes, due to typical illuminant colors, whereas outdoor light can bounce off diffuse objects with enough strength to create colorful reflections of whole scenes that can be blue in white balance due to the outdoor illuminant color; see \cref{fig:supp:dataset_overview}, examples $4$, $8$, $10$, $12$, $15$, $16$, $17$, $18$, $21$. At dusk, whole indoor scenes can reflect over cityscapes, etc.\ (examples $3$, $13$). Such priors are apparent in consumer photos (see Figs.~\ref{fig:main_result}, \ref{fig:gt_upcompare}, \ref{fig:results_usecases}, \ref{fig:supp:results_basecompare}, \ref{fig:supp:results_usecases}, \ref{fig:supp:results_usecases_more}). Lastly, like prior works we pair indoor/outdoor photos, which permits pairings such as bathrooms and beaches. Such pairs can be removed if they prove unhelpful.

\subsection{Base reflection removal}\label{sec:base_reflection_removal}

Base models were trained end-to-end from random weights at $256^2$ pixels using an Adam optimizer with $l_r=1e$-$4$, discriminator $l_r=5e$-$5$, and batch size $32$ over $16$ GPUs for $20$ epochs. Adversarial training begins after one epoch.

We trained three base models, one with and two without context $c$. To omit $c$, we removed the modulated merges (\cref{fig:base_model}), which decreases model capacity. As a second option, we left the model unchanged, and trained/tested with random $c$. We used this second approach for ablation.

\def\tplus{\raisebox{0.1em}{\tiny +}}
\def\btplus{\raisebox{0.1em}{\tiny {\bf +}}}
\def\tminus{\raisebox{0.05em}{\tiny $-$}}

\def\edgegap{\hspace{0.5em}}
\def\colgapa{\hspace{0.65em}}
\def\colgapb{\hspace{0.85em}}
\def\colgapc{\hspace{0.65em}}
\renewcommand{\arraystretch}{0.8}
\begin{table*}[t]
\begin{minipage}[t]{0.24933452\linewidth}    
  \vspace{-0.5em}
  \fontsize{7}{10}\selectfont
  \centering
  \begin{tabular}[t]{@{\edgegap}l@{\hspace{0.4em}}l@{\colgapa}c@{\colgapa}c@{\colgapa}c@{\colgapa}c@{\edgegap}}
    \toprule
    \multicolumn{2}{c}{Method}  & $\mathrm{SSIM}_\mathrm{t}$ & \%$\uparrow$ & $\mathrm{SSIM}_\mathrm{r}$ & \%$\uparrow$ \\
    \midrule
    & Control & $85.8$ &  & $49.0$ & \\
    & Full Sim. & $\mathbf{95.7}$ & $\mathbf{70}$ & $\mathbf{88.2}$ & $\mathbf{77}$ \\
    \midrule
    \multirow{8}{0.85em}{\rotatebox{90}{Ablate}} & (G)amma & $88.8$ & $21$ & $77.0$ & $55$ \\
    & Exposure & $94.2$ & $59$ & $86.1$ & $73$ \\
    & Pose & $95.0$ & $65$ & $85.4$ & $71$ \\
    & Fresnel & $95.3$ & $67$ & $87.3$ & $75$ \\
    & WB & $95.5$ & $68$ & $87.6$ & $76$ \\
    & Blur & $95.5$ & $68$ & $87.9$ & $76$ \\
    \addlinespace[0.1em]
    \cline{2-6}
    \addlinespace[0.3em]
    & All \raisebox{0.05em}{\tiny $-$} G & $90.2$ & $31$ & $71.3$ & $44$ \\
    & All \raisebox{0.075em}{\tiny $+$} G & $89.2$ & $24$ & $58.5$ & $19$ \\
    \bottomrule
  \end{tabular}
  \vspace{-0.5em}
  \caption{Ablated datasets were created, $1.2$M examples each. Model \emph{ours+ctx} was trained on these, and tested on the full-simulation test set. \%$\uparrow$ is \wrt control (see \cref{tab:base_ablation}). SSIM values are shown as percentages.}
  \label{tab:sim_ablation}
  \vspace{0.5em}
\end{minipage}
\hfill
\begin{minipage}[t]{0.4108252\linewidth}
  \vspace{-0.5em}
  \fontsize{7}{10}\selectfont
  \centering
   \begin{tabular}[t]{@{\edgegap}l@{\hspace{0.4em}}l@{\colgapb}c@{}c@{\colgapb}c@{\colgapb}c@{\hspace{1em}}c@{\colgapb}c@{\edgegap}}
    \toprule
    & Method & $\mathrm{PSNR}_\mathrm{t}$ & $\mathrm{SSIM}_\mathrm{t}$ & \%$\uparrow$ & $\mathrm{PSNR}_\mathrm{r}$ & $\mathrm{SSIM}_\mathrm{r}$ & \%$\uparrow$ \\
    \midrule
    & Control & $21.74$ & $85.8$ & & $12.48$ & $49.0$ & \\
    \multirow{5}{1em}{\rotatebox{90}{RAW Train\hspace{0.3em}}} & Ours{\tiny +}ctx & $\mathbf{33.23}$ & $\mathbf{95.7}$ & \btplus $\mathbf{70}$ & $\mathbf{30.17}$ & $\mathbf{88.2}$ & \btplus $\mathbf{77}$ \\
    & Ours & $32.15$ & $95.2$ & \tplus $66$ & $29.18$ & $86.7$ & \tplus $74$  \\
    & Zhu~\cite{zhu2023} & $29.84$ & $92.8$ & \tplus $49$ &  &  &  \\
    & DSRNet~\cite{qiming2023} & $28.98$ & $92.6$ & \tplus $48$ & $23.99$ & $75.5$ & \tplus $52$  \\
    & Zhang~\cite{zhang2018} & $26.23$ & $89.9$ & \tplus $28$ & $22.78$ & $61.5$ & \tplus $25$ \\
    & CoRRN~\cite{wan2020} & $22.75$ & $86.7$ & \tplus $6$ & $18.31$ & $60.4$ & \tplus $22$ \\
    \midrule
    \multirow{5}{1em}{\rotatebox{90}{$8$-bit Pub.}} & Control & $18.62$ & $78.4$ & & $9.79$ & $37.4$ &  \\
	& DSRNet~\cite{qiming2023} & $19.99$ & $80.0$ & \tplus $8$ & $16.98$ & $49.3$ & \tplus $19$ \\
    & Zhu~\cite{zhu2023} & $19.84$ & $79.7$ & \tplus  $6$ &  &  &  \\
    & Zhang~\cite{zhang2018} & $18.65$ & $75.9$ & \tminus $11$ & $17.37$ & $51.0$ & \tplus $22$ \\
    & CoRRN~\cite{wan2020} & $18.95$ & $74.7$ & \tminus $17$ & $15.99$ & $23.8$ & \tminus $22$ \\
    \midrule
    \multirow{2}{1em}{\rotatebox{90}{Abl.\hspace{0.25em}}} %
    & Ours{\tiny +}rac & $33.20$ & $95.7$ & \tplus $70$ & $30.20$ & $88.3$ & \tplus $77$ \\    
    & Ours{\tiny +}rnd & $32.42$ & $95.1$ & \tplus $66$ & $29.29$ & $86.7$ & \tplus $74$ \\
    \bottomrule
  \end{tabular}
  \vspace{-0.5em}
  \caption{Base models: (\emph{control}) compares $m$ to $t$, $r$. $8$-bit models use published weights. \%$\uparrow$ is \wrt SSIM control. Ablations:  (\emph{rac}) GT $r$ is used as context; (\emph{rnd}) random $c$. \emph{Ours+ctx} beats \emph{Ours+rnd} {\footnotesize ($p < 1.7e$-$11$)}.
  }
  \label{tab:base_ablation}
  \vspace{0.5em}
\end{minipage}
\hfill
\begin{minipage}[t]{0.30434783\linewidth}    
  \vspace{-0.5em}
  \fontsize{7}{10}\selectfont
  \centering
  \begin{tabular}[t]{@{\edgegap}l@{\hspace{0.4em}}l@{\hspace{0.2em}}c@{\colgapc}c@{\colgapc}c@{\colgapc}c@{\hspace{0.25em}}}
    \toprule
    & Method & PSNR$_\mathrm{t}$ & SSIM$_\mathrm{t}$ & PSNR$_\mathrm{r}$ & SSIM$_\mathrm{r}$ \\
    \midrule
    & Control & $19.50$ & $86.3$ & $13.13$ & $64.5$ \\
    \multirow{8}{0.85em}{\rotatebox{90}{GT}} & Ours & $\mathbf{47.77}$ & $\mathbf{98.8}$ & $\mathbf{45.93}$ & $\mathbf{98.2}$ \\
    & Ours-NM & $43.29$ & $98.1$ & $43.99$ & $96.9$ \\
    & VDSR+C~\cite{prasad2021}& $42.24$ & $97.9$ & $38.32$ & $93.8$ \\
    & VDSR~\cite{prasad2021} & $40.74$ & $97.4$ & $38.30$ & $93.9$ \\
    \cmidrule(r){2-6}
    & Bicubic & $31.98$ & $85.1$ & $41.58$ & $96.0$ \\
    & SUPIR~\cite{yu2024} & $28.09$ & $64.6$ & $28.29$ & $56.8$ \\
    & RESRGan~\cite{wang2021} & $23.72$ & $65.6$ & $23.11$ & $53.6$ \\
    \midrule
    \multirow{2}{0.85em}{\rotatebox{90}{E2E\hspace{0.25em}}}
    & Ours & $\mathbf{30.62}$ & $\mathbf{95.2}$ & $\mathbf{28.53}$ & $\mathbf{90.7}$ \\
    & VDSR+C~\cite{prasad2021} & $30.27$ & $94.5$ & $27.74$ & $88.7$ \\
    \bottomrule
  \end{tabular}
  \vspace{-0.5em}
  \caption{Upsampling ground truth (GT), and using the base model for end-to-end results (E2E). Usampling is from $256$p to $2048$p using our method and V-DESIRR with and without cycle consistency (+C), which improves VDSR for $\mathrm{T}$ {\footnotesize ($p$\hspace{0.2em}$<$\hspace{0.2em}$1e$-$12$)}.
  }
  \label{tab:upsampler_comparison}
  \vspace{0.5em}
\end{minipage}
\end{table*}

Our system uses RAW images end-to-end, but public datasets do not provide RAW images: Real20, Real45, Nature, SIR2, SIR2$^+$, CDR,\footnote{The authors of CDR \cite{lei2021b} have not released the RAW data.} and RRW all use JPG/PNG formats~\cite{zhang2018, fan2017, li2019, wan2017, wan2022, lei2021b, zhu2023}. We tabulate results using our simulation test sets, and show visual results using RAW photos that were captured in-the-wild. See also \cref{sec:supp:base_model_comparisons}.

In \cref{tab:base_ablation} we compare to Zhang \etal~\cite{zhang2018}, DSRNet~\cite{qiming2023}, Zhu \etal~\cite{zhu2023}, and CoRRN~\cite{wan2020} by retraining their models on our RAW dataset.\footnote{We use $2.5$M parameters; DSRNet uses $125$M. Inference at $256\times 341$ takes $0.96\mathrm{s}$/$1.04\mathrm{s}$ on a $2021$ M1 MacBook Pro ($32$Gb) and iPhone $14$ Pro.} Recall that our model uses the same losses and network head as Zhang \etal. This simplifies comparison to prior work. \cref{tab:base_ablation} (\emph{RAW Train}) shows that, when training with RAW, all methods improve images relative to the average degradation to $t,r$ (\emph{control}). Our models however outperform prior works ({\em ours+ctx}, {\em ours}). 

To show the benefit of RAW simulation and inference, we ran the previously published $8$-bit models on an 8-bit version of our test set,\footnote{Our test images were converted to 8-bit using Adobe Camera RAW.} and compared the percent improvement to that of using RAW (\cref{tab:base_ablation}, \emph{8-bit Pub.}). DSRNet and Zhu improve images modestly; Zhang and CoRRN distort the color \mbox{(\eg \cref{fig:main_result}).} {\bf Retraining DSRNet, Zhu, and Zhang on RAW improves their performance by $\approx$40 pct.\ points (pp) SSIM$_\textrm{t}$, whereas the performance differences among them are only $\approx$20pp. Training on RAW simulation data therefore improved performance more than the architectural variations among prior works. Furthermore, ablating the simulation (\cref{tab:sim_ablation}, \emph{All+G}) degrades performance \mbox{-46pp}, which conversely matches the +40pp benefit of RAW retraining, and exceeds even the benefit of the contextual photo (+4pp).}

In \cref{tab:base_ablation} (\emph{Abl.}) we ablate our contextual model by training/testing with random $c$ (\emph{ours+rnd}), which degrades performance compared to (\emph{ours+ctx})---this is statistically significant. Removing operations that use $c$ (\emph{ours}) did not degrade performance compared to \emph{ours+rnd} ($p < 1.7e$-$11$), which suggests that \emph{ours+rnd} does not learn dataset priors with its additional capacity, and conversely that \emph{ours+ctx} leverages the content of $c$. Ablating further, using the reflection as the context ($c = r$) \mbox{\em at test time only} does not improve the contextual model results (\emph{ours+rac}), which suggests that $c$ and $r$ need not match; the model is robust to their differences since it is trained with disjoint crops ($c$, $r$). 

For visual comparison, in \cref{fig:gt_upcompare} and \cref{fig:supp:results_basecompare} we captured\footnote{We thank Florian Kainz for his help capturing these photos.} ground truth reflections in common cases: looking outdoors, into a display case, and at artwork. We dereflected with Zhang, DSRNet (retrained, {\em RT}), and our models at $256\hspace{-0.15em}\times\hspace{-0.15em}384$ (inset images) and upsampled to $2048\hspace{-0.15em}\times\hspace{-0.15em}2731$ (next section). The empirical SSIM values (lowercase t, r) are commensurate with test performance (\cref{tab:base_ablation}). In \cref{fig:gt_upcompare} our contextual model separates the reflection, but without context our model attributes the colors in the umbrella with a reflected object. Prior works perform quantitatively worse.

In Figs.\ \ref{fig:main_result}, \ref{fig:extra_result}, \ref{fig:results_usecases}, \ref{fig:supp:results_usecases}, \ref{fig:supp:results_usecases_more}, and \ref{fig:supp:extra_result} we show results on photos in-the-wild from cameras that were not used to construct the training data. We also compare the $8$-bit models of Zhang and DSRNet. The bottom two rows of Figs.~\ref{fig:results_usecases}, \ref{fig:supp:results_usecases}, \ref{fig:supp:results_usecases_more} show that these prior $8$-bit models perform qualitatively worse than when they are re-trained/evaluated on RAW (the top rows). They do not however recover $r$ well, which is needed for aesthetics and error correction (\cref{sec:editing}, \cref{sec:supp:editing}).

\textbf{Discussion.} Our models recover $t,r$ in diverse real-world cases including museums, nature, shopping, a mid-day city, artwork, etc.\ (Figs.~\ref{fig:main_result}, \ref{fig:extra_result}, \ref{fig:gt_upcompare}, \ref{fig:results_usecases}, \ref{fig:supp:results_basecompare}, \ref{fig:supp:results_usecases}, \ref{fig:supp:results_usecases_more}, \ref{fig:supp:extra_result}). In \cref{fig:main_result}, using the context photo yields more correct and uniform color on the Egyptian tablet because there is less ambiguity about the color of the reflection scene (compare to inset {\em w/o context}). Failures occur when $t$ or $r$ is bright, and pushes the other into the noise floor, saturating it to black---the problem becomes hole filling. When a single color channel saturates, the content can sometimes be recovered. But, systems must address hole filling because users typically cannot control the strength of reflections.

Errors can occur when textured regions of $t$ and $r$ overlap, as in \cref{fig:main_result} where a stone wall overlaps the subject's dress. Color differences help: in \cref{fig:gt_upcompare} the reflected painting is separated from the tree. Without such differences, models must repair or hallucinate content in the corrupted $t$. Saturated reflections pose a similar challenge. See \cref{sec:supp:results} for more discussion of errors and additional results.

\subsection{Upsampling}\label{sec:upsampling_reflection_removal}

Our upsampler is trained using Adam with $l_r=4e$-$4$, batch size $64$ over $32$ $\mathrm{A}100$ GPUs, and converges after about $40$ epochs. For end-to-end operation (E2E), we tune with the base model outputs for $19\mathrm{K}$ examples at $l_r=2e$-$4$.

We compare to V-DESIRR~\cite{prasad2021} in \cref{tab:upsampler_comparison} by upsampling the ground truth (GT) and using the base model (E2E).\footnote{Inference of our E2E upsampler, up to preview size $1024\times 1364$, takes $4.52\mathrm{s}$ and $6.53\mathrm{s}$ on our 2021 MacBook Prop and iPhone 14 Pro.} For best E2E performance, we fine tuned our upsampler and V-DESIRR with the base model. Our method performs best ({\em ours}). Cycle consistency loss improves V-DESIRR (\emph{+C}), so we used this for E2E. We ablated the upsampler masking operations by using only the finisher head ({\em Ours-NM}); performance degraded almost to match V-DESIRR.

Comparing on GT images, \cref{fig:results_upcompare} and \cref{fig:supp:results_upcompare}, V-DESIRR produces strong artifacts, even after fine tuning (adding cycle-consistency losses did not help).

\textbf{Discussion.} V-DESIRR amplifies errors at low resolutions by repeatedly upsampling its previous output images. Instead, our model masks and copies the high-res mixture features $\phi_M$ to the output $T, R$. This direct copy reduces error propagation. Errors can still occur when features that are not present in the low resolution inputs become visible at the next level upward (\eg \cref{fig:supp:upsampler_failure}) because the low resolution $t, r$ cannot guide upsampling of such features, and the %
\begin{wrapfigure}[13]{r}{0.45\linewidth}
  \centering
    \vspace{-0.75em}
    \includegraphics[width=1\linewidth]{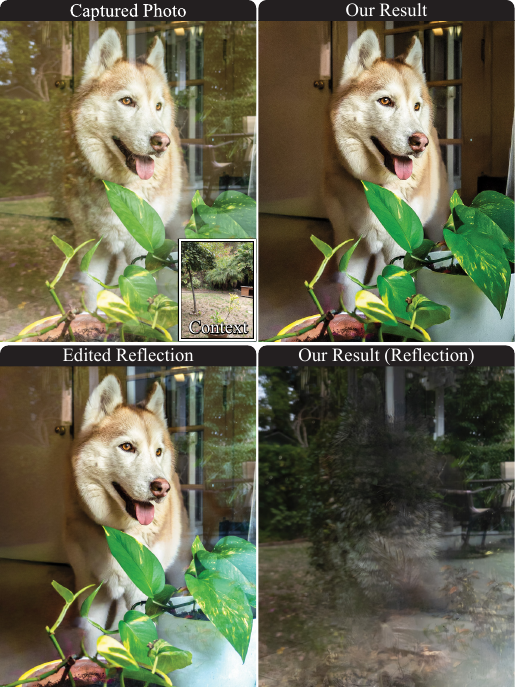}   
    \vspace{-1.5em}
    \caption{Reflection editing.}
    \label{fig:editing}
    \vspace{0.75em}
\end{wrapfigure}%
upsampler must infer the high resolution image to which the features belong. 
\subsection{Reflection editing}\label{sec:editing}

In \cref{fig:editing} and \cref{fig:supp:editing_repair} we show that the predicted reflection facilitates aesthetic editing and error correction. In \cref{fig:editing}, the reflection color and spatial arrangement is modified. Error correction is shown in \cref{fig:supp:editing_repair}, \cref{sec:supp:editing}. Edits were made in Photoshop using the tone-mapped $t$ and $r$ images, and ``Linear Dodge'' blend mode (but linear blending would be ideal).

\begin{figure*}[h!]
    \begin{minipage}[b]{0.5818138599880524\linewidth}    
        \centering
        \includegraphics[width=\linewidth]{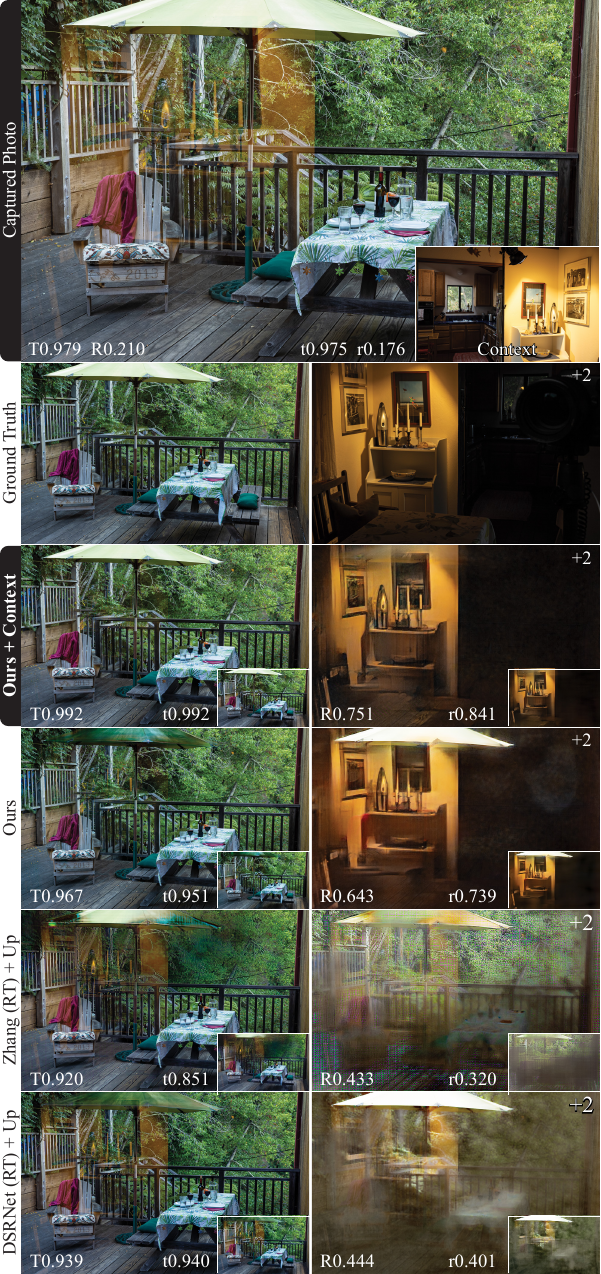}
        \caption{Comparisons to ground truth (GT) at $256\times 384$ (inset) and $2048\times 3072$. Methods are retrained for RAW (RT). SSIMs are relative to GT.
        }
        \label{fig:gt_upcompare}
    \end{minipage}
    \hfill
    \begin{minipage}[b]{0.3737407620880983\linewidth}    
        \centering
        \includegraphics[width=\linewidth]{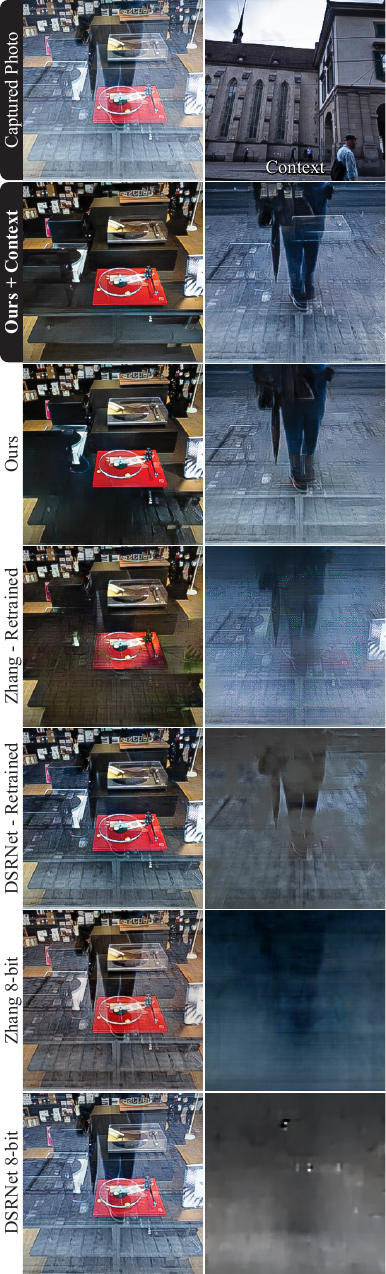}
        \caption{Comparisons to models trained on 8-bit images (bottom), with results at $256^2$ pixels.
        }
        \label{fig:results_usecases}
    \end{minipage}
    \vspace{-2em}
\end{figure*}

\section{Conclusion}
\label{sec:conclusion}

We have described a de-reflection system that is trained solely on images from a photometrically and geometrically accurate simulation. Moreover, we have imbued these images with natural priors by searching among millions of them for well-exposed and visually interpretable cases. This RAW simulation dramatically improves results, more than prior model variations, and enables our models to perform well on real images without training on them.

Since Farid and Adelson~\cite{farid1999}, many cues have been used for de-reflection. We add illuminant color and context photos, and use RAW images end-to-end. Our models are thus sensitive and can uncover hidden reflections, \cref{fig:supp:extra_result}; privacy should be protected. Our system can also remove lens flares, though they are not in the dataset, \cref{fig:supp:flare_result}. Flare removal systems might therefore be pre-trained to remove reflections, since it is difficult to capture real lens flares.

{
    \small
    \bibliographystyle{ieeenat_fullname}
    \bibliography{main}
}

\clearpage
\setcounter{page}{1}
\maketitlesupplementary

\def\algofontsize#1{\fontsize{8}{10}\selectfont}

\begin{appendices}
\setcounter{figure}{0}
\setcounter{algorithm}{0}
\renewcommand\thefigure{S\arabic{figure}}   
\renewcommand\thealgorithm{S\arabic{algorithm}}

\section*{Contents}\label{sec:supp:outline}

{\bf Supplemental sections}
\begin{enumerate}[label={}, leftmargin=5.5em, labelwidth=4.5em, labelsep=0.3em, align=left]
    \item[\Cref{sec:supp:photometric_sim}.] {\em Photometric reflection synthesis}
	\begin{enumerate}[label={}, leftmargin=-1.5em, labelwidth=2em, labelsep=0.3em, align=left]
	    \item[\ref{sec:supp:white_balancing}.] White balancing
	\end{enumerate}
	\vspace{0.125em}
    \item[\Cref{sec:supp:geometric_sim}.] {\em Geometric reflection synthesis}
	\begin{enumerate}[label={}, leftmargin=-1.5em, labelwidth=2em, labelsep=0.3em, align=left]
	    \item[\ref{sec:supp:fresnel_attenuation}.] Fresnel attenuation
	    \item[\ref{sec:supp:camera_projection}.] Camera projection
	    \item[\ref{sec:supp:defocus_blur}.] Defocus blur
	    \item[\ref{sec:supp:hdr_env_sampling}.] HDR environment sampling
	    \item[\ref{sec:supp:double_reflection}.] Double reflection
	\end{enumerate}
	\vspace{0.125em}
    \item[\Cref{sec:supp:contextual_photo}.] {\em Contextual photos}
    \item[\Cref{sec:supp:data_collection}.] {\em Data collection}
	\begin{enumerate}[label={}, leftmargin=-1.5em, labelwidth=2em, labelsep=0.3em, align=left]
	    \item[\ref{sec:supp:mixture_search}.] Mixture search
	    \item[\ref{sec:supp:image_collection}.] Image collection
	    \item[\ref{sec:supp:dataset_settings}.] Dataset settings
	\end{enumerate}
	\vspace{0.125em}
    \item[\Cref{sec:supp:reflection_removal}.] {\em Reflection removal methods}
	\begin{enumerate}[label={}, leftmargin=-1.5em, labelwidth=2em, labelsep=0.3em, align=left]    
	    \item[\ref{sec:supp:base_model}.] Base model architecture
	    \item[\ref{sec:supp:upsampler}.] Upsampler architecture
	\end{enumerate}
	\vspace{0.125em}
    \item[\Cref{sec:supp:results}.] {\em Results}
	\begin{enumerate}[label={}, leftmargin=-1.5em, labelwidth=2em, labelsep=0.3em, align=left]
	    \item[\ref{sec:supp:evaluation_methods}.] Evaluation methodology
	    \item[\ref{sec:supp:base_model_comparisons}.] Base model comparisons
	    \item[\ref{sec:supp:upsampler_comparisons}.] Upsampler comparisons
	    \item[\ref{sec:supp:editing}.] Editing applications
	\end{enumerate}
	\vspace{0.125em}
    \item[\Cref{supp:sec:adobe_camera_raw}.] {\em Adobe Camera RAW DNG SDK}
\end{enumerate}

\noindent {\bf Supplemental figures}
\begin{enumerate}[label={}, leftmargin=5em, labelwidth=4em, labelsep=0.3em, align=left]
    \item[\cref{fig:dataset_geometric}.] Examples of simulated geometric properties
    \item[\cref{fig:supp:dataset_overview}.] Overview of the reflection dataset
    \item[\cref{fig:supp:selfie_geometry}.] Contextual camera geometry
    \item[\cref{fig:base_model}.] Base model architecture
    \item[\cref{fig:upsampler}.] Upsampler architecture
    \item[\cref{fig:supp:results_basecompare}.] Result comparisons with ground truth
    \item[\cref{fig:supp:results_usecases}.] Base model results in-the-wild
    \item[\cref{fig:supp:results_usecases_more}.] Base model results in-the-wild (continued)
    \item[\cref{fig:supp:extra_result}.] Results on subtle reflections
    \item[\cref{fig:supp:flare_result}.] Results for lens flare removal
    \item[\cref{fig:supp:results_upcompare}.] Upsampler model comparison
    \item[\cref{fig:supp:upsampler_failure}.] Upsampler model failures
    \item[\cref{fig:supp:editing_repair}.] Reflection editing for error recovery
\end{enumerate}

\noindent {\bf Supplemental functions (simulation and DNG SDK)}
\begin{enumerate}[label={}, leftmargin=5.2em, labelwidth=4.2em, labelsep=0.3em, align=left]
    \item[\fref{func:compute_exposure}.] \texttt{\small compute\_exposure()}
    \item[\fref{func:white_balance}.] \texttt{\small white\_balance()}
    \item[\fref{func:extract_xyz_images}.] \texttt{\small extract\_xyz\_images()}
    \item[\fref{func:cam_to_rgb}.] \texttt{\small cam\_to\_rgb()}
    \item[\fref{func:stage3_black_level}.] \texttt{\small stage3\_black\_level()}
    \item[\fref{func:highlight_recovery}.] \texttt{\small highlight\_recovery()}
    \item[\fref{func:find_xyz_to_cam}.] \texttt{\small find\_xyz\_to\_cam()}
    \item[\fref{func:xyz_to_cam_awb}.] \texttt{\small xyz\_to\_cam\_awb()}
    \item[\fref{func:calc_whitexy} - \ref{func:srgb_to_linear_srgb}.] Additional ACR functions
\end{enumerate}

\section{Photometric reflection synthesis}\label{sec:supp:photometric_sim}

As part of our photometric reflection synthesis pipeline, \fref{func:simulate_examples}, we compute a new exposure and white balance for the simulated mixture image, $m$, using \mbox{\fref{func:compute_exposure}} and \fref{func:white_balance}, respectively. These functions follow ACR color processing, and use methods in the Adobe DNG SDK, \cref{supp:sec:adobe_camera_raw}. The ACR color processing that produces XYZ source images is specified in \mbox{\fref{func:extract_xyz_images}} and discussed in \cref{supp:sec:adobe_camera_raw}. 

\subsection{White balancing}\label{sec:supp:white_balancing}

To compute a new white balance within \fref{func:white_balance}, we use the C5 white balancer of Afifi \etal \cite{afifi2020}. C5 white balances an input image by using an additional $n$ sample images that were captured from the same camera. We therefore cache $n$ samples for each camera in the dataset of RAW images, and remove all images for which there were not $n=7$ samples from the camera ($7$ is the C5 default).

\begin{algorithm}[b!]
\algofontsize{}
\caption{Compute the exposure of a simulated mixture $m$.}
\footnotesize
\textbf{Input:} A simulated mixture $m$ and its associated component images $(t, r)$ \\
\textbf{Output:} An exposure value $e$
\begin{algorithmic}[1]
    \STATE Compute the \tt{WhiteXY} of $t$. \hfill\COMMENT{SDK \fref{func:calc_whitexy}}%
    \STATE Compute transform \tt{XYZ\_to\_sRGB} using \tt{WhiteXY}. \hfill\COMMENT{SDK \fref{func:XYZ_to_linear_sRGB}}%
	\STATE Convert $m$ to linear sRGB using \tt{XYZ\_to\_sRGB}.
    \IF{no pixels in $t$ or $r$ are saturated}
    		\STATE Compute the mean pixel value $\mu$ of $m$
		\STATE Compute the target value $\tau =$ \tt{sRGB\_to\_linear\_sRGB(}$0.4$\tt{)}. \hfill\COMMENT{SDK \fref{func:srgb_to_linear_srgb}}%
    		\RETURN $e$ = $\tau / \mu$ \hfill\COMMENT{Expose the mean to sRGB $0.4$.}
	\ELSE
		\STATE Convert $t$ and $r$ to linear sRGB using \tt{XYZ\_to\_sRGB}.
		\STATE Compute $t_\mathrm{max} = \max(t)$ and $r_\mathrm{max} = \max(r)$.
		\RETURN $e = 1 / \min(t_\mathrm{max}, r_\mathrm{max})$
    \ENDIF
\end{algorithmic}
\label{func:compute_exposure}
\end{algorithm}

White balancing with C5 requires that simulated mixtures $m = t + r$ in XYZ be transformed into camera color space. In \fref{func:white_balance} we use the \ttbig{XYZ\_to\_CAM} transform associated with the RAW source image of $t$ to simulate a camera from which the mixture was captured, since $t$ typically dominates $r$ in the sum due to attenuation by the glass (see \cref{sec:supp:geometric_sim}). The white balancer produces a new %
\begin{algorithm}[t!]
\algofontsize{}
\caption{Compute the white balance transform.}
\footnotesize
\textbf{Input:} A re-exposed XYZ mixture, $e\cdot m$, and the transmission $t$ \\
\textbf{Output:} \tt{XYZ\_to\_XYZ\_awb}
\begin{algorithmic}[1]
	\STATE Compute the \tt{XYZ\_to\_CAM} using the \tt{WhiteXY} of $t$. \hfill\COMMENT{SDK \fref{func:find_xyz_to_cam}}
    \STATE Transform $e\cdot m$ into camera color space using \tt{XYZ\_to\_CAM}.
    \STATE Compute a new white point \tt{WhiteCAM\_awb} in camera color space. \hfill\COMMENT{This work uses \cite{afifi2020}}
    \STATE Compute \tt{XYZ\_to\_CAM\_awb} and \tt{WhiteXY\_awb} from \tt{WhiteCAM\_awb}. \hfill\COMMENT{\fref{func:neutral_to_xy}+\ref{func:find_xyz_to_cam} or \ref{func:xyz_to_cam_awb}}\label{func:white_balance:line_get_white_xy}
    \STATE Compute \tt{XYZ\_to\_XYZ\_D50} from \tt{WhiteXY\_awb}. \hfill\COMMENT{SDK \fref{func:map_white_matrix}}\label{func:white_balance:bradford}
    \RETURN \tt{XYZ\_to\_XYZ\_D50} $\boldsymbol{\cdot}$ \tt{inv(XYZ\_to\_CAM\_awb)} $\boldsymbol{\cdot}$  \tt{XYZ\_to\_CAM}.\label{func:white_balance:xyz_to_xyz_awb}
\end{algorithmic}
\label{func:white_balance}
\end{algorithm}%
white point \ttbig{WhiteCAM\_awb} in camera color space. We then follow ACR color processing of white points in camera color space by transforming \ttbig{WhiteCAM\_awb} into XY coordinates and computing a new \ttbig{XYZ\_to\_CAM} transform using DNG SDK \mbox{\fref{func:find_xyz_to_cam}}. This new transform into XYZ is then composed with Bradford adaptation (\fref{func:white_balance} line~\ref{func:white_balance:bradford}) to construct a single, linear white balancing transform that operates on images in XYZ space (line~\ref{func:white_balance:xyz_to_xyz_awb}).

Over a large scale dataset, white balancer failures inevitably occur. These are handled by culling $m$ if the new white point is extremely different from the as-shot \ttbig{WhiteXY} of $t$. Extreme changes to the true white point are not common because reflections that are of practical interest are either transparent, localized, or both. The new XY white point (\ttbig{WhiteXY\_awb}) can be further restricted to lie on the Planckian locus, and we found this to be sufficient for our source images, which were captured under typical illuminants. Projection from \ttbig{WhiteCAM\_awb} to \ttbig{WhiteXY\_awb} can be done using ACR \fref{func:neutral_to_xy} and \ref{func:find_xyz_to_cam}, or \mbox{\fref{func:xyz_to_cam_awb}} with the Planckian constraint, as noted in \mbox{\fref{func:white_balance}}, line~\ref{func:white_balance:line_get_white_xy}.

In summary, the white balance computed in \fref{func:white_balance} is a linear transform, which we denote \ttbig{XYZ\_to\_XYZ\_awb}, that composes three ACR color transforms: 1) it maps images into the camera color space of $t$, 2) it maps back to XYZ under a new white point, and 3) it applies Bradford adaptation to the D50 illuminant. In the simulation (\fref{func:simulate_examples}) this transform is applied to $m$, $t$, $r$, and $c$ so they are interpreted with respect to the same white point, lines \ref{func:simulate_examples:calc_wb}-\ref{func:simulate_examples:wb_trc}.

\section{Geometric reflection synthesis}\label{sec:supp:geometric_sim}

As part of our reflection removal pipeline, a geometric simulation is used to construct transmission and reflection pairs of images \mb{(t, r)} from a dataset of pairs of scene-referred (RAW) photographs \mb{(i,j) \in \mcD} that were not captured with or through glass. These transmission and reflection pairs are then added together to form the training data for our models, with ground truth provided by the constituent images in each pair. This simulation approach overcomes the scaling bottleneck of capturing real reflection images for training, which is difficult because ground truth (without the glass present) is not readily available.

\begin{figure}[t!]
  \centering
   \includegraphics[width=\linewidth]{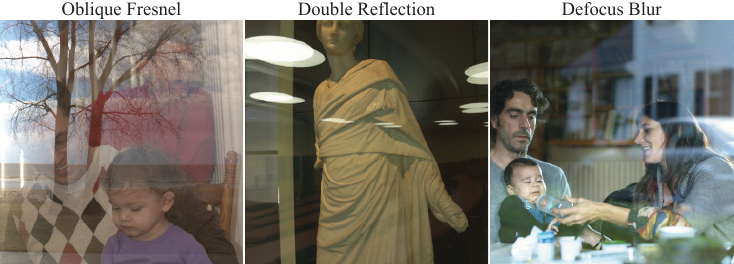}   
   \caption{Simulated geometric properties at extreme values. Blurs are typically subtle.}
   \label{fig:dataset_geometric}
\end{figure}

In particular, we synthesize transmission images \mbox{$t = T(i)$} and reflection images \mbox{$r = R(j)$} as functions of $i$ and $j$ that appropriately model Fresnel attenuation, perspective projection, double reflection, and defocus blur. We omit from $T$ effects related to global color, dirt, and scratches since existing photo editing tools are well equipped to correct them after reflection removal. Examples are shown in \cref{fig:dataset_geometric}, and an overview of the synthetic images is shown in \cref{fig:supp:dataset_overview}. 

\subsection{Fresnel attenuation}\label{sec:supp:fresnel_attenuation}

Fresnel attenuation is the most essential property to simulate because it reduces the intensity of the reflected image. Specifically, reflections $r$ are attenuated by a spatially varying factor $\alpha$ that depends on the angle of incidence $\theta_\mathrm{i}$ at which light strikes the glass with respect to the surface normal vector. As derived in~\cite{kong2014},
\begin{align}
	\alpha & =
	\frac{1}{2}(\alpha_{\perp} + \alpha_{\parallel})
\\
    \alpha_{\perp} & =
    \frac{
        \sin^2 (\theta_\mathrm{i} - \theta_\mathrm{i}')
    }{
        \sin^2 (\theta_\mathrm{i} + \theta_\mathrm{i}')
    }
\\
    \alpha_{\parallel} & =
    \frac{
        \tan^2 (\theta_\mathrm{i} - \theta_\mathrm{i}')
    }{
        \tan^2 (\theta_\mathrm{i} + \theta_\mathrm{i}')
    }~~,
\end{align}
where \mbox{$\theta_\mathrm{i}' = \arcsin(\frac{1}{\kappa}\sin \theta_\mathrm{i})$}, and $\kappa$ is the refractive index of glass. For \mbox{$\theta_\mathrm{i} \in [0^\circ$, $45^\circ]$}, Fresnel attenuation accounts for up to $-4$ stops (underexposure), and gradually strengthens to $-1$ stop for rays that glancingly strike the glass at $83^\circ$.

To specify $\theta_\mathrm{i}$, we define images \mb{r = R(j)} as originating from a mirror surface, with incident rays reaching the camera by the law of reflection. In the next section we decribe how to simulate a diversity of practical geometric configurations of the glass and camera to construct $\theta_\mathrm{i}$ and thus compute $\alpha$. Glass also attenuates the transmission \mb{t = T(i)} by \mb{1 - \alpha}. This is typically close to $1$, but at extreme angles it creates a visible darkening effect (\cref{fig:supp:dataset_overview}, example $26$). Typical attenuation levels are shown in \cref{fig:supp:dataset_overview} in the column labeled ``reflection.''

\subsection{Camera projection}\label{sec:supp:camera_projection}

We model consumer photography applications in which one sees a subject partially visible behind glass and takes a picture of it. This constrains the relative pose of the camera and glass, and introduces natural priors on the location and appearance of reflections. For example, skies typically reflect near the top of images, and reflections are typically stronger at the edges of photos where the camera rays strike the glass at a relatively higher angle of incidence, $\theta_\mathrm{i}$.

\begin{figure*}[h!]
    \centering
    \includegraphics[width=0.825\linewidth]{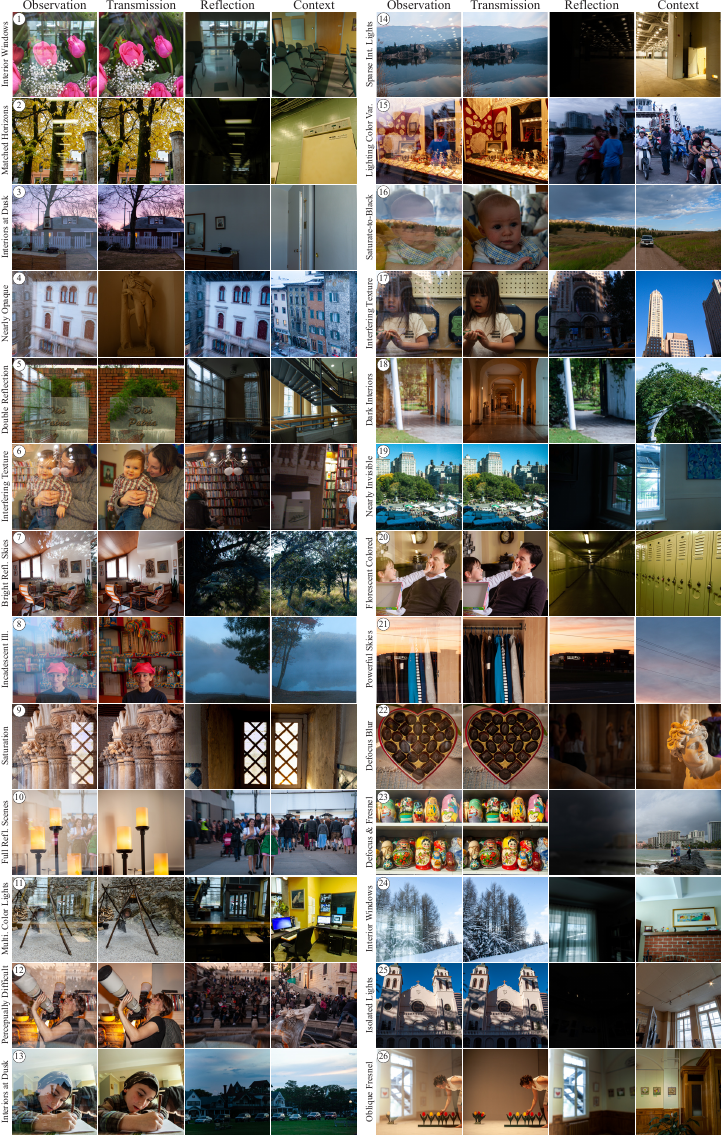}   
    \vspace{-0.5em}
    \caption{Dataset overview ($2048$p). {\bf All images are simulations that use RAW sources.} The numbers are referenced in the text.}
    \label{fig:supp:dataset_overview}
    \vspace{-4.5em}
\end{figure*} %
\textbf{Inclination angle $\phi$}. Most glass is approximately vertical, so if the viewpoint of $t$ looks upward, the viewpoint of $r$ should as well. We use a pose estimator~\cite{holdgeoffroy2017}, and augment the search for realistic pairings $(t, r)$, \cref{sec:photometric_synthesis}, by checking if \mb{\phi_t - \phi_r = \Delta_\phi} is below a maximum absolute value. In addition to this inclination discrepancy filter, images are culled if their inclination angle $\phi$ exceeds a threshold. This aligns the horizons of $t$ and $r$, which introduces spatial priors, as illustrated in \cref{fig:supp:dataset_overview} examples $2$, $5$, $14$, $20$, and $24$.

\textbf{Roll angle $\rho$}. Images are culled if their estimated roll $\rho$ exceeds a maximum absolute value as these typically indicate that the pose estimator has failed. 

\mbox{\textbf{Field of view}}. Images are also culled if the estimated vertical FOV is zero, which indicates general pose estimation failures. Otherwise we randomly sample \mb{\mathrm{FOV} \sim \mathcal{U}(\mathrm{FOV}_\mathrm{min}, \mathrm{FOV}_\mathrm{max})}, where $\mathcal{U}$ is the uniform distribution.

\textbf{Azimuth angle $\theta$}. Most glass in consumer photography is roughly planar. We constrain the camera azimuth with respect to the glass so that the camera rays strike this plane (accounting for the FOV). We randomly sample $\theta \sim \mathcal{U}(-\theta_\mathrm{max}, \theta_\mathrm{max})$. The effect of this constraint can be most easily seen in \cref{fig:supp:dataset_overview} where highly oblique camera angles create spatially varying Fresnel attenuation across the reflection component (examples $6$, $7$, $14$, $19$, $23$, and $26$).

\subsection{Defocus blur}\label{sec:supp:defocus_blur}

Recently Lei~\cite{lei2021b} found that performance of state-of-the-art methods degrades significantly for sharp reflections due to an imbalance of blurry images in training and testing data. Physically based methods have been developed to introduce realistic defocus blur using depth maps~\cite{kim2019}, but this introduces a data collection bottleneck by requiring RGBD cameras that also have physical limitations. We instead model a physically based prior on the amount of defocus blur.

Defocus blurs are determined by the camera focus depth, aperture, and focal length. Points on an object at depth $d_o$ that differs from the focus depth $d_f$ project to a circular region with diameter $\delta$,
\begin{align}
    \delta & =
    \frac{|d_o - d_f|}{d_o}
    \frac{f^2}{N(d_f - f)}~~,
\end{align}
where $f$ is the focal length, and $N$ is the aperture f-number. This {\em circle of confusion}~\cite{goldberg1992} is magnified with increasing focal length $f$ or decreasing N.

Defocused images are simulated by sampling diameters $\delta$ (mm) for the circle of confusion. The focal length $f$ (mm) and aperture $N$ (dimensionless) are sampled according to their physical ranges in mobile cameras. The object and focus depths $d_o$ and $d_f$ (feet) are sampled \mb{d \sim \mathcal{U}(d_{\min}, d_{\max})}, in the plausible and finite range of scene depths to which $\delta$ is sensitive. The diameter $\delta$ (mm) is converted to a percentage of the sensor height, $\delta_p$ (the minimum sensor dimension). Reflections \mb{r = R(j)} are blurred by convolving them with a circular defocus kernel with pixel diameter $h\delta_p$, where $h$ is the minimum dimension of the image $j$ in pixels. We maintain this physical calibration when images are cropped into halves to simulate contextual views (\cref{sec:contextual_photo} and \Cref{sec:supp:contextual_photo}). 

Our physically based sampling procedure simulates reflections with a realistic amount of defocus blur for consumer photography. An example of a strongly blurred reflection is shown in \cref{fig:dataset_geometric}. We however find that reflections are typically sharp, as Lei~\cite{lei2021b} also notes. Typical defocus blurs are shown in \cref{fig:supp:dataset_overview}; strong blurs are shown in examples $22$ and $23$.

\subsection{HDR Environment sampling}\label{sec:supp:hdr_env_sampling}

A dataset of indoor $360^\circ$ HDR Image-Based-Lights (IBLs) are used as an additional source of scene-referred images~\cite{gardner2017}. Artificial light sources in HDR images are typically not saturated, which makes it possible to simulate reflections of light sources that are not saturated (or, under-exposed RAW images could be used).

When one of the images in a pair \mb{(i, j) \in \mcD} is an IBL, a synthetic camera is constructed with a pose that matches the RAW image to which it is paired (see \cref{sec:supp:camera_projection}), excepting that the azimuth $\theta$ is sampled independently and uniformly at random in $360^\circ$. Contextual images $c$ are simulated by a second synthetic camera within the IBL with an adjacent, non-overlapping FOV.

The IBLs~\cite{gardner2017} are captured under a fixed white point, which allows for the color of the illuminant (i.e., its white balance) to be mixed correctly with the RAW data. We calibrate the exposure of these indoor IBLs by setting their median intensity to match the median value of all indoor RAW images (the median contends with saturated pixels). This cropped HDR image can be photometrically combined, and geometrically transformed using functions $T$ or $R$. The effect of HDRs is shown in \cref{fig:supp:dataset_overview}: reflected light sources, windows, etc.\ are produced by HDRs in examples $1$, $2$, $5$, $6$, $11$, $14$, $20$, $24$, and $26$.

\subsection{Double reflection}\label{sec:supp:double_reflection}

Glass panes introduce multiple reflective surfaces that create a double reflection or ``ghosting'' effect. \mbox{Shih et al.~\cite{shih2015}} ascribe the effect of double reflection to the thickness of a single or double pane, and show shifts up to $4$ pixels for thicknesses in 3--10mm under some viewing distances, but double reflections are often much larger. Gaps between panes reach $20$mm as reported commercially, and each pane adds up to $7$mm. These multiple reflecting surfaces are also not necessarily parallel, uniformly thick, or flat as assumed in~\cite{shih2015}. These factors produce significant double reflections even in modern windows, including when the camera is distant. We simulate these complex effects by adopting the geometric model of~\cite{shih2015} and allowing a greater range of thicknesses, \mbox{8--20mm}. We uniformly sample a glass thickness, physical viewing distance, and refractive index. These facilitate a ray tracing procedure, detailed below. \cref{fig:supp:dataset_overview} shows double reflections in the dataset; see examples $2$, $4$, $5$, $7$, $8$, $12$, $14$, and $15$. 

The primary reflection that contributes to \mb{r = R(j)} is determined by the Fresnel attenuation $\alpha j$ as described in \cref{sec:supp:fresnel_attenuation}. Specifically, the intensity of light at each homogeneous image coordinate $\m x$ is \mb{\alpha(\m x) j(\m x)}, because we have defined $j(\m x)$ as encoding the light along the incident rays $\m r$ with $\angle (\m x, \m r) = 2 \theta_\mathrm{i}$ where $\theta_\mathrm{i}$ is the angle of incidence. We simulate a second reflection by tracing the camera rays $\m x$ through a simulated single pane of uniform thickness to identify the coordinates $\m x'$ at which they would emerge from the glass after being internally reflected from the back surface of the pane. Coordinates $\m x'$ are shifted according to their transit distance within the glass, which is determined by the law of reflection and Snell's law. We neglect the latter as insignificant in comparison to the former and the various non-modeled physical effects of real glass panes. Under these assumptions, rays that enter the glass at $\m x$ emerge at $\m x'$ in direction $\m r$. An image $j'$ is needed to describe the intensity of incident light in direction $\m r$ at $\m x'$, but $j'(\m x') \ne j(\m x')$ because we have defined $j(\m x')$ as describing the light reflected from a corresponding direction $\m r'$. Since we do not have $j'$, we assume that the light field is sufficiently smooth that \mb{j'(\m x') \approx j(\m x')}, since \mb{\angle(\m r, \m r') = \angle (\m x, \m x')} is small. We therefore warp $j$ such that $\m x'\mapsto \m x$, and combine this warped image $j_w$ with $j$ to produce a double reflection image.

The double reflection image is given by $j_d = \alpha j + \beta j_w$, where $\alpha$ is the known Fresnel attenuation due to the primary reflection (see \cref{sec:supp:fresnel_attenuation}), and $\beta$ specifies the attenuation of the rays that travel into the glass before they are internally reflected back to the camera. These latter rays encounter three surfaces, and lose intensity at each one. The first surface is the front face of the glass, where they are mildly attenuated by \mb{1 - \alpha} as they transmit into the glass. Second is the back face, where they reflect and are attenuated again according to their angle of incidence, which has been altered by Snell's law. This change of incidence angle however has a negligible effect on the the Fresnel attenuation factor within the typical incidence ranges. We therefore use $\alpha$ as the attenuation at the second surface. Lastly, the rays re-encounter the front face of the glass (now from within) where they transmit out of the glass and are attenuated again by approximately \mb{1 - \alpha}. This gives \mb{\beta = (1 -\alpha)\alpha(1 -\alpha)}.

\cref{fig:dataset_geometric} shows an example of a simulated double reflection, selected to show a case when the primary and secondary reflections are significantly shifted. We note that no doubling effect can occur along the direction of the glass surface normal because the rays that enter the glass re-emerge after internal reflection at the same location they entered. Thus double reflection fields that follow our geometric model (and the model of~\cite{shih2015}), in which there are two perfectly parallel planes, must exhibit a radial pattern around the  image of the glass surface normal. These patterns are not always apparent in practice, which suggests that the geometrical arrangement of glass surfaces that is described by Shih~\cite{shih2015} omits important factors. We nonetheless adopt their model as being sufficient because visible reflections are typically localized to regions of an image, which obscures the presence or absence of a radial center. 

\section{The contextual photo}\label{sec:supp:contextual_photo}

\begin{figure}[h!]
    \centering
   \includegraphics[width=0.8\linewidth]{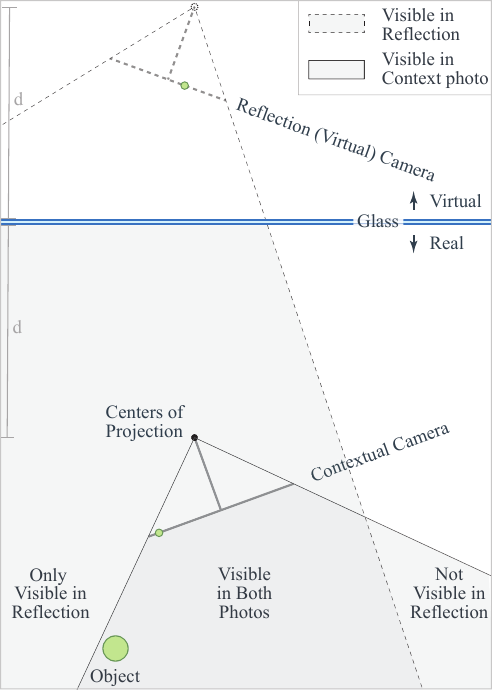}
   \caption{The geometry of a primary and contextual camera view.  In this figure the two views are co-located (black dot), and the latter is rotated 180$^\circ$ with respect to the former. Neither condition is required or typical; they are shown to illustrate one possible geometric arrangement. The contextual camera frustum is shown at the bottom ($\perp$ symbol); the primary camera frustum is not drawn. The reflection contains the scenery that would be captured by a virtual camera behind the glass (open circle, dashed $\perp$), equidistant to the glass wrt the primary camera, and swung azimuthally in the opposite direction. Note, the object (green circle) appears at the right edge of the contextual camera's image (small green circle above it), but slightly left of center in the virtual camera view (small green circle near top of figure), and hence it is slightly right of center in the captured primary photo because the virtual camera is flipped left/right.
   }
   \label{fig:supp:selfie_geometry}
\end{figure}

One arrangement of a primary and a contextual camera is shown in \cref{fig:supp:selfie_geometry} (see caption for explanation). This specific arrangement of cameras is neither required nor typical in practice, but it reveals general geometric differences between the views of primary and contextual cameras. The view of the so-called reflection camera is translated by $2d$, twice the distance $d$ to the glass. Furthermore, if the contextual camera is rotated $180^\circ$ from the primary, the latter view will be in an opposite direction. At extreme rotations, the views will have little or no overlap. 

Because the translation and rotation of the contextual camera view can differ %
significantly from the primary camera, it is difficult to simulate a contextual view using a dataset of image pairs \mb{t=T(i)} and \mb{r=R(j)} that are used to create mixture images $m$ from the perspective of a primary camera. In particular, content from $j$ should not be copied into a simulated contextual image $c$, as trained models could learn to cheat by searching for patches of $r$ that have the same perspective projection in $c$. Such patches will not be present in practice. 

We create a scalably large dataset of contextual images $c$ by noting that $c$ will often contain no common content with $r$ unless the photographer is asked to point the camera at what they see in the reflection. We minimize such burden on the photographer, and define the contextual image $c$ as any image of the reflection scene that does not view the same parts of the scene as $r$. This definition allows contextual images to capture lighting information (sunlight, incandescent, etc.) and scene semantics (outdoor, indoor, city, nature, etc.) to aid reflection removal.

To construct $c$, we crop reflection source images, $j$, into non-overlapping left/right or top/bottom squares $(j_0, j_1)$, and similarly for transmission source images $i$. This yields four pairs for simulation $(i_a, j_b) : (a,b) \in \{0, 1\}^2$ for all $(i,j) \in \mcD$. The mixture and context images are $(m_{ab}, c_{1-b})$, where $m_{ab} = \mathcal{C}(T(i_a) + R(j_b))$, and $c_{1-b} = \mathcal{C}(j_{1-b})$. We fix the capture function $\mathcal{C}$ for both $m$ and $c$ to have the same white balance so $c$ can describe the color of the reflection scene in addition to its semantics. \cref{fig:supp:dataset_overview} shows example contextual photos.

\section{Data collection}\label{sec:supp:data_collection}

Below are the data search and collection methods summarized in \cref{sec:photometric_synthesis}.

\subsection{Mixture search}\label{sec:supp:mixture_search}

Well-exposed mixtures $m$ are identified by checking if the mean pixel value is within a normal distribution over the pixel values in the dataset of RAW images. 

Well mixed $m$ are identified by computing the SSIM between $m$ and $t$ as a block-wise image, and checking if the mean of this SSIM image is within a useful range: if the SSIM is too high, the reflection is imperceptible; if it is too low, the mixture is not visually interpretable, even by a human. We compute this single channel SSIM image as a weighted average of the corresponding per-channel SSIM images. The weights are the average value in each color channel, which better accounts for strongly colored images. Lastly, the standard deviation of this single channel SSIM image is checked to remove reflections that are imperceptible, but nonetheless produce a low mean SSIM by spreading their power broadly (they have low spatial variance). The effect of this search can be seen in \cref{fig:supp:dataset_overview}: simulated reflections can be faint or nearly invisible (examples $19$, $22$, $23$, and $25$); or so strong that the transmission is nearly invisible or slightly difficult interpret (examples $4$, $12$, $16$, and $21$).

\subsection{Source images}\label{sec:supp:image_collection}

We collect all images at their native RAW camera resolution to facilitate training upsampling methods. We label all images, including IBLs (\cref{sec:supp:hdr_env_sampling}), as outdoor $\mcO$ and indoor $\mcI$ since glass typically separates indoor and outdoor spaces. This information is available in existing datasets~\cite{mit5k, raise, gardner2017}, and can be collected at large scales via crowd-sourcing. We define our dataset $\mcD$ of pairs \mb{(i, j)} for simulation as \mb{\mcD = (\mcO\times \mcI) ~\cup~ (\mcI\times \mcO) ~\cup~ (\mcI\times \mcI) - \mcP}, where $\mcP$ is all pairs \mb{i=j}. The set \mb{\mcO \times \mcO} is uncommon, and should be included sparingly following empirical priors. We omit them. 

These pairings of source images interact with the mixture search process (\cref{sec:supp:mixture_search}) to introduce photometric and semantic priors, as seen in \cref{fig:supp:dataset_overview}. Outdoor transmission scenes typically have reflections of indoor light sources or windows (examples $1$, $2$, $5$, $9$, $11$, $14$, $19$, $24$) unless it is dusk (example $3$). Indoor transmission scenes typically have strong reflections of skies or full outdoor scenes due to the brightness of natural light (examples $4$, $7$, $8$, $10$, $12$, $15$, $16$, $17$, $18$, $21$) unless it is dusk (example $13$). Lastly, indoor-indoor pairings are often complex because $t$ and $r$ are typically similar in brightness (examples $6$, $20$, $26$).

\begin{figure*}[h!]
  \centering
   \includegraphics[width=\linewidth]{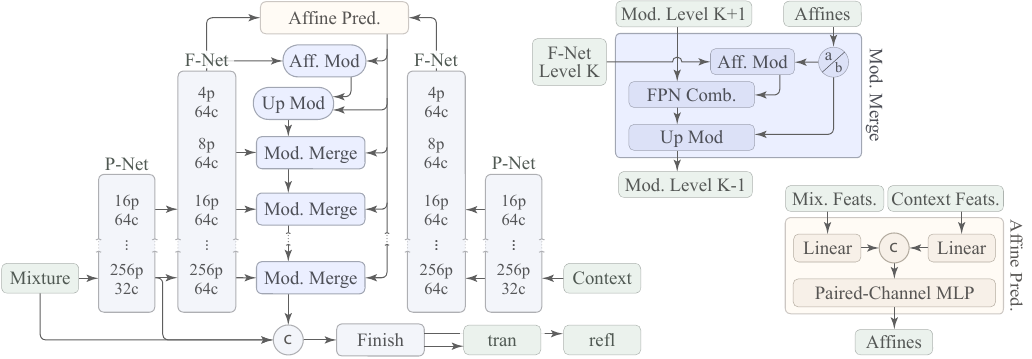}
   \caption{The base model. Mixture and context images are projected into a high dimensional space using a shared backbone~\cite{effnet} (labeled {\em P-Net}; weights are shared), and a feature fusion network~\cite{effdet} (labeled {\em F-Net}; weights are shared). The context features are used to predict affine transforms for each feature channel at each resolution. Channel-wise modulation is used because contextual photos do not always include content that can be matched. Modulation can help identify the reflection in the feature space. We use two conv-mod-deconv operations of~\cite{karras2020} within the modulated merge blocks. The FPN combine op is a fast normalized fusion module from the BiFPN architecture~\cite{effdet}. 
   }
   \label{fig:base_model}
\end{figure*}

\subsection{Simulation settings}\label{sec:supp:dataset_settings}

Two {\em capture scenarios} are generated for each pair \mb{(i, j) \in \mathcal{D}}. A virtual camera is posed randomly with maximum azimuth \mb{\theta_\mathrm{max} = 50^\circ} toward the glass and \mb{\mathrm{FOV} \sim \mathcal{U}(50^\circ, 80^\circ)}, where $\mathcal{U}$ denotes the uniform distribution. Pairs $(i, j)$ are culled if $|\Delta_\phi| > 15^\circ$, or either image has absolute inclination value $|\phi| > 45^\circ$ or roll  $|\rho| > 10^\circ$ (see \cref{sec:supp:camera_projection}). Lastly, capture scenarios are also culled if the camera rays from more than $4$ pixels do not strike the glass (they are parallel or divergent). This final check ensures that the glass fills the FOV.

We compute spatially varying Fresnel attenuation with index of refraction \mb{\kappa \sim \mathcal{U}(1.47, 1.53)}~\cite{glassindex}; see \cref{sec:supp:fresnel_attenuation}. Double reflections are simulated with glass thickness (mm) in $\mathcal{U}(8, 20)$ at distances (mm) in $\mathcal{U}(500, 2000)$, with $50\%$ probability of being a double pane; see \cref{sec:supp:double_reflection}. Defocus blur is simulated with object and focus distances (ft) in $\mathcal{U}(1, 100)$ with aperture and focal length of iPhone main cameras, $N\approx1.6$ and $f\approx 26$mm ($35$mm equivalent units); see \cref{sec:supp:defocus_blur}. Simulated mixtures are culled if the mean SSIM between $m$ and $t$ falls outside of \mb{[0.4, 0.94]}, or if the standard deviation of this SSIM is below $0.05$; see \cref{sec:supp:mixture_search}.

\section{Reflection removal}\label{sec:supp:reflection_removal}

Here we provide details of the base model architecture, as summarized in \cref{sec:base_model}, and the upsampler, \cref{sec:upsampler}.

\subsection{Base model}\label{sec:supp:base_model}

The base model is designed to leverage local and global features, \cref{fig:base_model}, and produce $256^2$ pixel outputs in about $1$ second on a mobile device to meet req.\ \ref{req:fast_review} (see \cref{sec:introduction}).

A feature backbone~\cite{effnet} is used to project $m$ into a linear, high dimensional space ($32$-D) and compute semantic features (labeled {\em P-Net}). Features are at a variety of spatial and channel resolutions: $(256, 32)$, $(256, 16)$, $(128, 24)$, $(64, 40)$, $(32, 112)$. These features include the outputs of the initial convolution layer of the \ttbig{EfficientNet-B1} variant of \cite{effnet} (as implemented by \cite{timm}), which we modify to use an initial stride of $1$ rather than $2$ so no initial down-sampling is performed on the input $256\times 256$ pixel images. 

The multi-resolution feature tensors from the backbone are next fused into $64$ channels at the input resolution using the \ttbig{D0} variant of the EfficientDet feature pyramid architecture~\cite{effdet, timm} (labeled {\em F-Net}). This architecture first augments the input features with three additional levels: $(16, 64)$, $(8, 64)$, $(4, 64)$, where each results from a $2\times2$ maxpool with stride $2$, and the first is preceded by a $1\times 1$ conv, batch norm, and no activation. The augmented input features are then input to a series of so-called BiFPN layers~\cite{effdet} (see Fig.~3), which fuse features from low resolution to high, and then back to low resolution, in a zigzag operation that is repeated three times. To obtain high resolution fused features at only the input resolution of $256\mathrm{p}$, we add a fourth repetition in which we omit the final high-to-low pass. We furthermore modify the low-to-high pass to incorporate the contextual image, as described next.

The contextual image, $c$, is passed through the same F-Net and P-Net using the same weights as $m$. The features of $c$ and $m$ at the lowest resolution are input to an affine prediction module, \cref{fig:base_model} (lower right). This module first vectorizes its two $64\times 4\times 4$ inputs, and passes them through a fully connected layer to transform them into two $64$-$\mathrm{D}$ vectors. These vectors embody $64$ pairs of channels, which we concatenate and input to an MLP (labeled {\em paired-channel MLP}) that predicts affine transforms that modify the features of the FPN during the final low-to-high pass.

The paired-channel MLP is a series of grouped convolutions that implement $64$ independent MLPs followed by a fully connected layer. These $64$ MLPs each have $2$ inputs, $2$ hidden, and $1$ output dimension, with leaky ReLUs after each layer. The inputs to these MLPs are corresponding pairs of channels from $m$ and $c$. The outputs compose a single $64$-$\mathrm{D}$ vector that is input to a fully connected layer to predict $64\times K\times 2$ affine transforms, two for each of the $64\times K$ channels and levels of the FPN.\footnote{Note that we include two levels at $256^2$ pixels, in correspondence with the resolutions of the features that we extract from the backbone.}  Conceptually, this paired-channel MLP has the capacity to compare $c$ and $m$ to identify channels that match the reflection scene, and to determine how to transform those channels to remove reflections.

The predicted affines from the paired-channel MLP are used in the final low-to-high pass of the FPN to perform a series of modulated merge operations, \cref{fig:base_model} (upper right). These merge operations use two affine transforms per feature channel, labeled $a$ and $b$ in \cref{fig:base_model}. Transforms $a$ are used to perform a conv-mod-deconv operation ala StyleGan~\cite{karras2020} on the features of $m$ from FPN level $K$. These features are subsequently combined with the features from level $K+1$ by resampling the latter features $2\times$ and using fast normalized fusion~\cite{effdet} (labeled {\em FPN Combine}). These combined features are modified with a second conv-mod-deconv operation using the second group of affines, $b$.

The final modulated merge produces $64$-$\mathrm{D}$ features at $256^2$ pixels. These features are concatenated with $m$ and the features from the first layer, for a total of \mb{3+32+64=99} channels. A convolutional finishing module is then applied. This module has the capacity to further identify and finally render $t$ and $r$. To simplify comparison to prior work, our finishing module is the head in~\cite{zhang2018}. The first layer is $1\times 1$, and projects the $99$ input features to $64$-D, which is maintained in the remaining operations. Those operations are $3\times 3$ convolutions dilated by $(1, 2, 4, 8, 16, 32, 64, 1)$; each is followed by a batch norm and leaky ReLU. A final $1\times 1$ convolution generates $6$ channels: $t$ and $r$.

The model is trained using the perceptual, adversarial, and gradient losses of Zhang \etal \cite{zhang2018} with a ResNet-based discriminator~\cite{karras2020}, and optimized 5-tap derivatives~\cite{farid2004} in the exclusion loss to suppress grid artifacts. We also adopt the $l_1$ reflection loss of~\cite{zhang2018} to minimize arbitrary differences to prior work. Perceptual losses are computed in non-linear sRGB by applying gamma compression and using VGG19 features, weighted to contribute equally. Crucially, we train end-to-end from randomly initialized weights.

\subsection{Upsampler}\label{sec:supp:upsampler}
\begin{figure}[b]
  \centering
   \includegraphics[width=\linewidth]{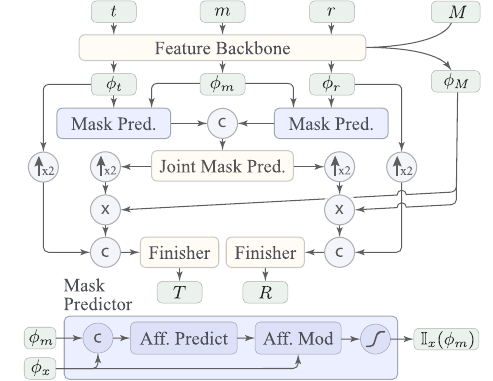}
   \caption{Upsampler at one pyramid level.}
   \label{fig:upsampler}
\end{figure}

The upsampler is shown in \cref{fig:upsampler} and introduced in \cref{sec:upsampler}. It transforms low resolution outputs $t$, $r$ from the base model to a flexible output resolution. We use a Gaussian pyramid and apply the same upsampler at each level. The upsampler projects the low-res, $3$-channel inputs $(m, r, t)$ into a high dimensional space $\phi$ using convolutions in an expand-and-contract pattern. We use $3\times3$ kernels for feature expansion, and $1\times 1$ kernels for contraction. There are $3$ layers with hidden dimensions $(32, 16), (64, 32), (128, 64)$. We use leaky ReLU between the hidden layers, and no skip connections. Batch norms are omitted to facilitate a feature matching process, described next.

The components $t$ and $r$ are separated by identifying low resolution features $\phi_t$ and $\phi_r$ in the low resolution mixture $\phi_m$. We predict $2$, per-pixel, per-channel low resolution masks using a mask prediction module, \cref{fig:upsampler} (bottom), which uses a paired-channel MLP (defined in \cref{sec:supp:base_model}) to predict its affine transforms (see also \cref{sec:upsampler}). The joint mask predictor also uses a paired-channel MLP, but it directly outputs the final masks rather than affine transforms. The input, hidden, and output dimensions of both paired-channel MLPs are $(2, 2, 2, 1, 2)$. The final layer is fully connected. As noted in \cref{sec:supp:base_model}, these MLPs can be implemented efficiently as a series of grouped $1\times 1$ convolutions. 

The finishing network is a series of $3\times 3$ convolutions that are distinguished by the number of channels and dilation rates, $128$:$(1, 2)$, $96$:$(1, 2, 4, 8)$, $64$:$(1, 1)$. A final $1\times 1$ convolution produces the $6$ channels of output for $T$ and $R$.

The upsampler is trained using a cycle-consistency loss on the predicted transmission and reflection, in addition to the losses of Pawan et al.~\cite{prasad2021}. Perceptual features are computed by converting to non-linear sRGB. For each predicted high resolution image \mb{\xhe \in \{\The, \rhe\}} the loss is a weighted sum of the following terms: \mb{\E[| \xhe - \xh |]}, $\E[(\xhe - \xh)^2]$, $\E[|\nabla \xhe - \nabla \xh|]$,  $\textsc{LPIPS}(\xhe, \xh)$ and $\E[|D(\xhe) - \xl |]$, where $D$ downsamples $\xhe$ to compute the cycle consistency loss, and $\E$ denotes expectation over spatial dimensions and (where applicable) the output of the gradient operator. We use both the $l_1$ and $l_2$ norms to avoid introducing arbitrary differences to prior work \cite{prasad2021}. The norms are weighted $0.2$, gradient terms $0.4$, LPIPS $0.8$, and cycle consistency $10$. These losses are accumulated over three upsampling levels from $128^2$ to $1024^2$ pixels (smaller than the $256^2$ pixel output size of the base model to contend with memory constraints during training). The upsampler is trained first on the ground truth low resolution inputs, and fine tuned on the output of the base de-reflection model. When fine tuning, the $256^2$ pixel outputs of the base model are downsampled to 128$^2$ pixels. At test time, no downsampling is applied. The upsampler takes $256^2$ pixel images as input, and produces output at $2048^2$ pixels and higher. 

\section{Results}\label{sec:supp:results}

Here we provide results and discussion in addition to \cref{sec:results}.

\subsection{Evaluation methods}\label{sec:supp:evaluation_methods}

\begin{figure*}[h!]
  \centering
    \includegraphics[width=\linewidth]{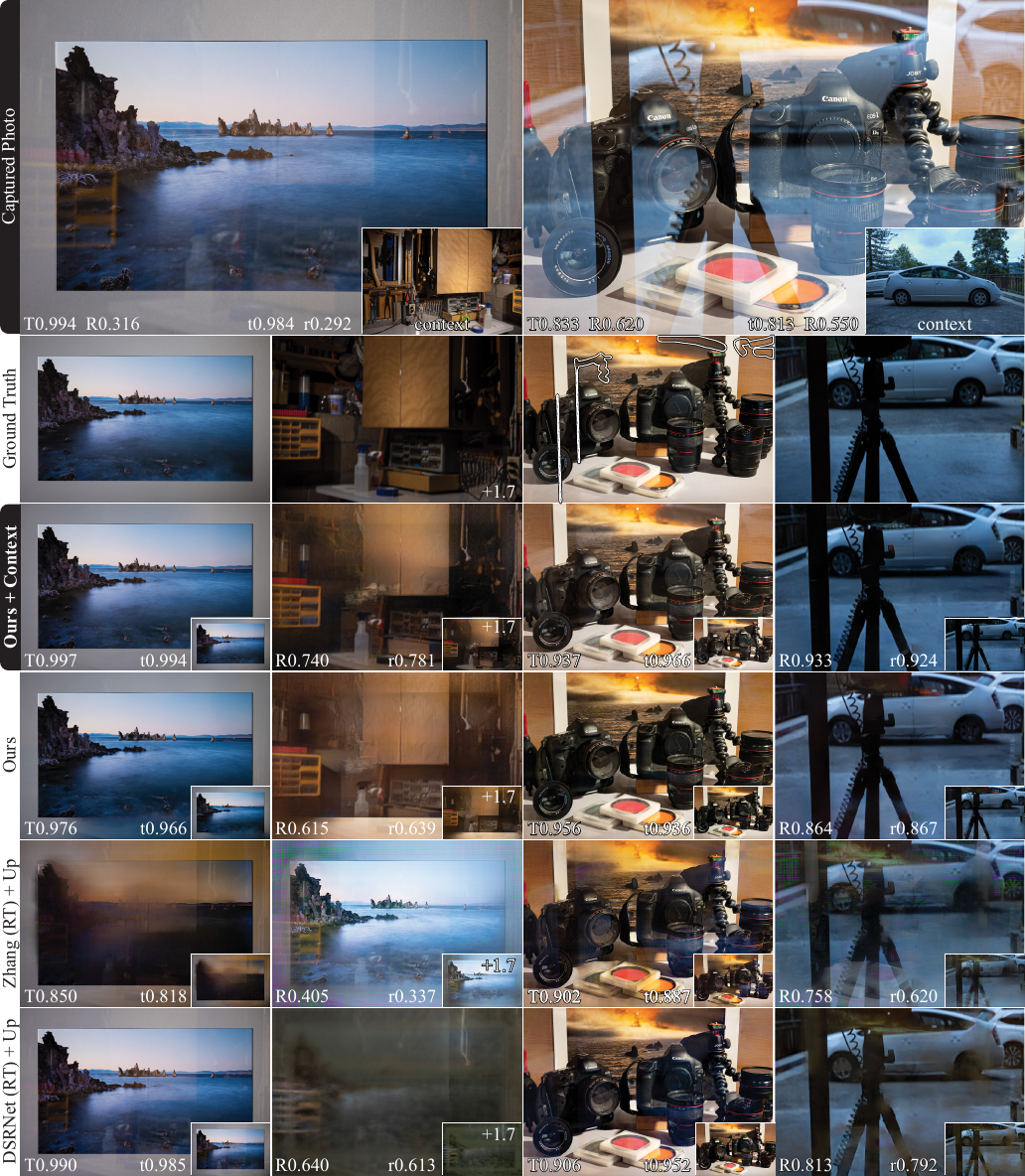}   
    \caption{Comparisons with ground truth at $256\times384$ and $2048\times3072$. The re-trained, low-resolution methods Zhang~\cite{zhang2018} and DSRNet~\cite{qiming2023} are upsampled using our upsampler, and both low- and high-resolution results are shown. The SSIM of the predicted $t$ and $r$ is reported at low resolution (labeled t, r) and high (labeled T, R). Errors in ground truth are outlined and omitted from the SSIM.
    }
    \label{fig:supp:results_basecompare}
\end{figure*}

\textbf{Ground truth capture}. Ground truth capture was used for \cref{fig:results_upcompare}, \cref{fig:gt_upcompare}, and \cref{fig:supp:results_basecompare}. Mixture images $m$ were captured with ground truth $r$ by placing a black material behind the glass and taking a second photo. Images $t$ were computed \mb{t = m - r} in linear sRGB after normalizing the exposures and using the white point of $m$ (see ACR Step \ref{acr:convert_rgb}, \cref{sec:reflection_synthesis}). Nonetheless, we found it necessary to capture $m$ and $r$ with fixed exposures and white points to avoid imprecisions in the values that are stored in RAW metadata.

\textbf{SSIM computation}. We report SSIM values between pairs of RAW images $(a, b)$ by first transforming them into linear sRGB (ACR Step \ref{acr:convert_rgb}, \cref{sec:reflection_synthesis}) using the white point of $a$. By using a consistent white balance, across the ground truth $m$, $t$, and $r$, we penalize errors in the white balance of the estimated $t$ and $r$. For SSIM computation, images are then converted to non-linear sRGB by applying standard gamma compression \mbox{(ACR Step \ref{acr:gamma_compression})}:
\begin{align}\nonumber
    x_\mathrm{sRGB} = 
    \begin{cases}
        12.92x & x \le 0.0031308  \\
        1.055 x^{1/2.4}  & x > 0.0031308
    \end{cases}    
\end{align}
where $x$ are pixel values in a linear sRGB image. We omit tone mapping operations (ACR Step \ref{acr:tone_mapping}) to remove subjectivity from the SSIM values. Lastly, SSIMs are reported as averages over the low-resolution images (denoted t, r) and high-resolution images (denoted T, R). 

\textbf{Ground truth photographs}. Ground truth images were captured in three common scenarios: 1) looking out of a home window, \cref{fig:gt_upcompare}; 2) photographing artwork, \cref{fig:supp:results_basecompare} (left); and 3) looking into a display case, \cref{fig:supp:results_basecompare} (right). In scenario one, the illuminant in the reflection scene is approximately $3$,$000\mathrm{K}$, and the white point of the mixture image is $7$,$300\mathrm{K}$. In scenario two, these values are $6$,$000\mathrm{K}$ and $6$,$850\mathrm{K}$, and in scenario three they are $6$,$000\mathrm{K}$ and $3$,$627\mathrm{K}$.

\subsection{Base model comparisons}\label{sec:supp:base_model_comparisons}

\textbf{\mbox{Reflections with ground truth}}. In \cref{fig:supp:results_basecompare} we show two images with ground truth reflections: photographing artwork, and looking into a display case. These complement \cref{fig:gt_upcompare}, in which a photo was taken when looking out of a home window. The home window view includes an interior reflection that is strongly yellow due to the indoor illuminant color. In contrast, the artwork in \cref{fig:supp:results_basecompare} is illuminated by the same light source as the reflection scene, which produces a correctly white balanced reflection that consequently has more diverse colors. In contrast, the display case in \cref{fig:supp:results_basecompare} (right) reflects an outdoor scene with a different white balance, which produces a strong, blue reflection. This outdoor reflection is visible over broad regions because the illuminant is powerful enough to reflect off of the diffuse ground and sidewalk surfaces with sufficient intensity to be visible over the contents of the display case. These exterior reflections are also qualitatively different from those in photo of the artwork, where the reflection is sparse. The SSIM of the artwork is therefore high on average ($0.994$), but the reflections are locally strong, whereas the SSIM of the display case is low ($0.833$) because the reflection affects broad regions.

Our base models improve the SSIM of $t$ and $r$ in all of these ground truth examples (labeled in lower case t, r), and this extends to the upsampled results (labeled in uppercase T, R) whereas prior works do not perform as well (\cref{fig:supp:results_basecompare} and \cref{fig:gt_upcompare}).  Our contextual model produces a more correct transmission and reflection image on the artwork. On the display case, the contextual model improves the reflection, whereas the cars are not fully removed. We believe this variation results from saturated regions in the sky of the contextual photo, where we use the as-shot illuminant color in the EXIF to recover the color of the saturated pixels. Lastly, the method of Zhang~\cite{zhang2018} associates blue colored content with the reflection in both images, but this is incorrect for the painting; the transmission image is therefore distorted.

\begin{figure*}[t!]
  \centering
    \includegraphics[width=\linewidth]{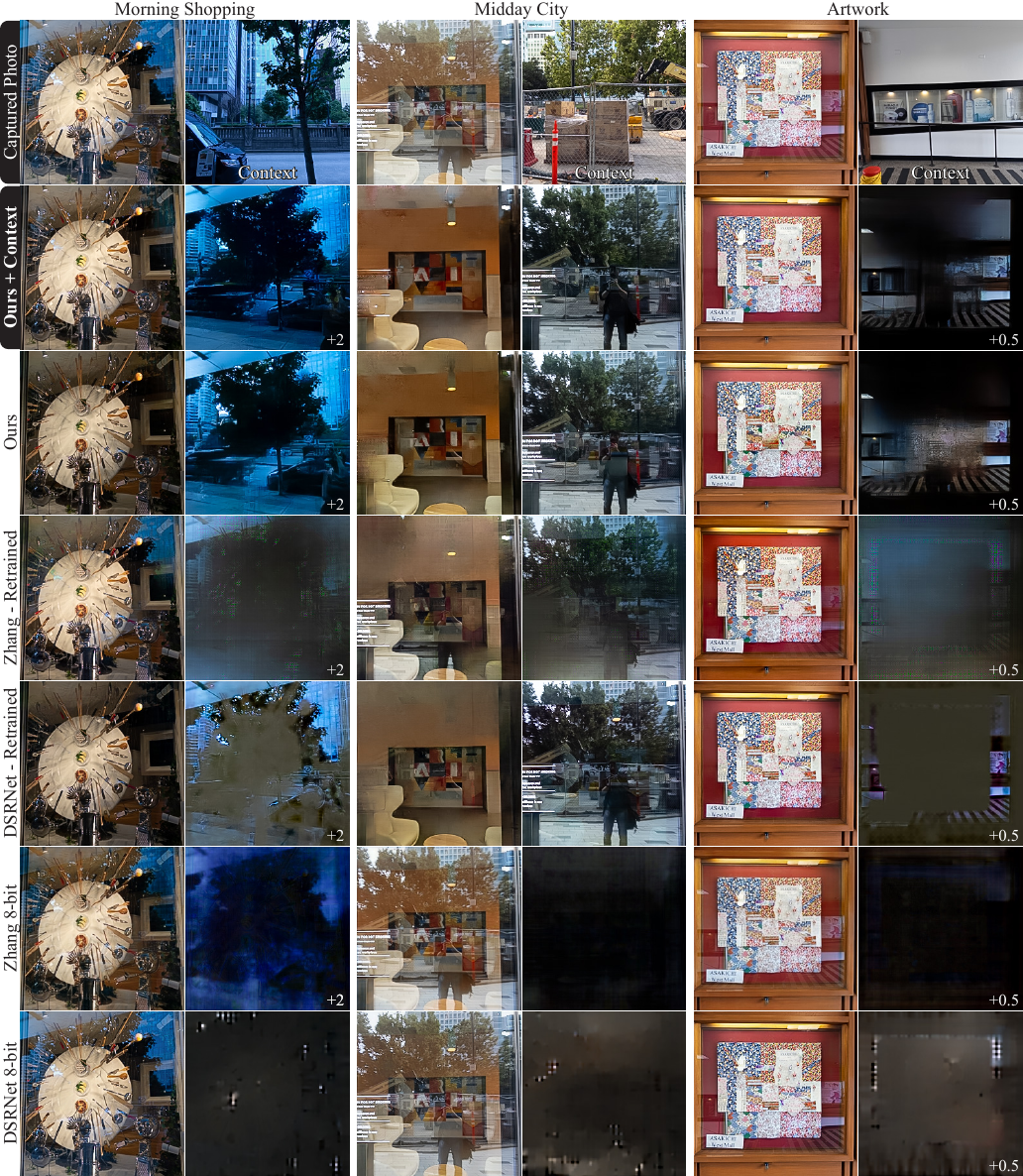}   
    \caption{Results for base models at $256^2$ pixels. Reflections marked “+X” are brightened by X stops compared to the transmission.
    }
    \label{fig:supp:results_usecases}
    \vspace{2em}
\end{figure*}

\begin{figure*}[t!]
  \centering
    \includegraphics[width=\linewidth]{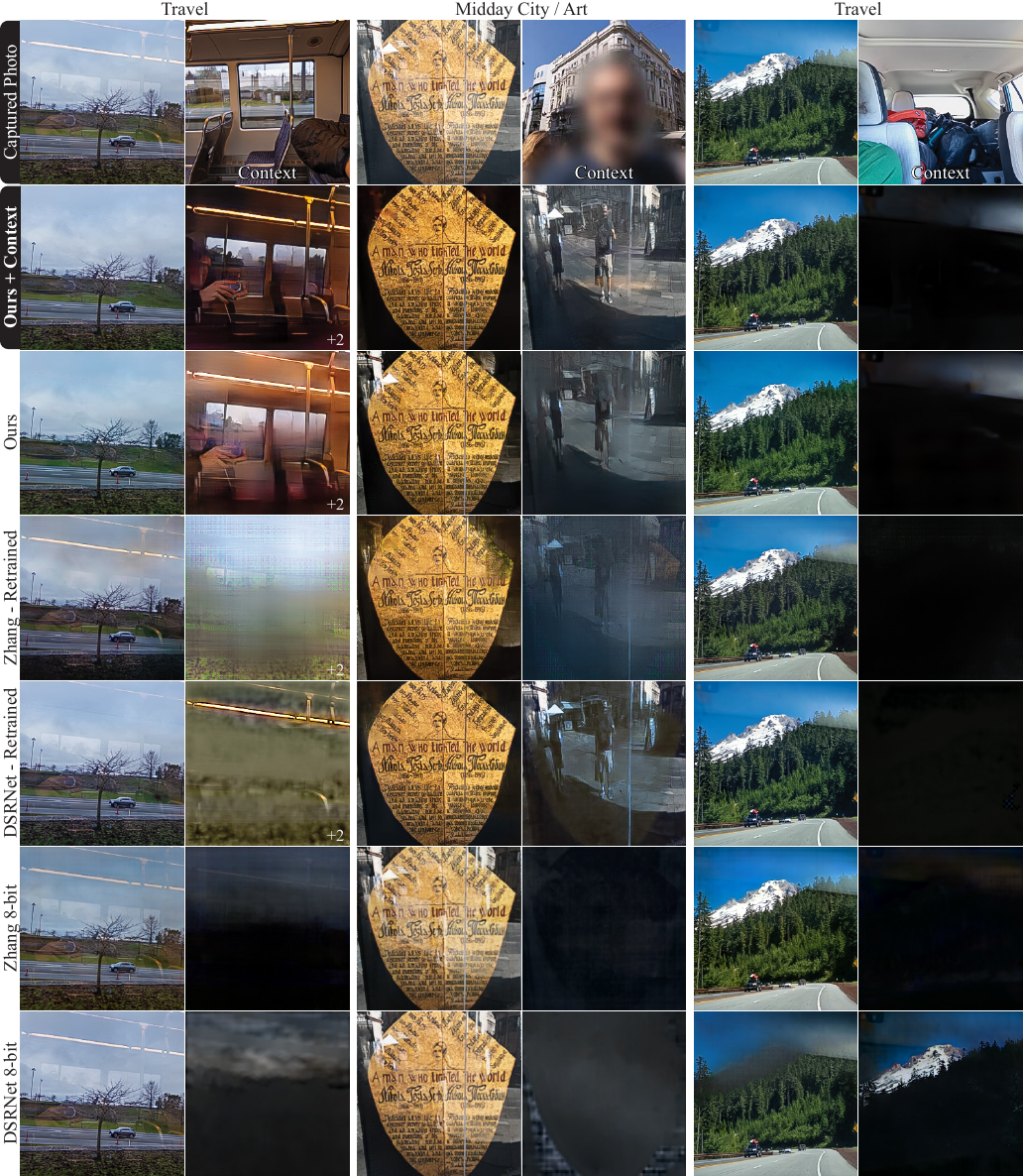}   
    \caption{Results for base models at $256^2$ pixels. Reflections marked “+X” are brightened by X stops compared to the transmission.
    }
    \label{fig:supp:results_usecases_more}
    \vspace{2em}
\end{figure*}%

\begin{figure*}[h!!]
  \centering
    \includegraphics[width=\linewidth]{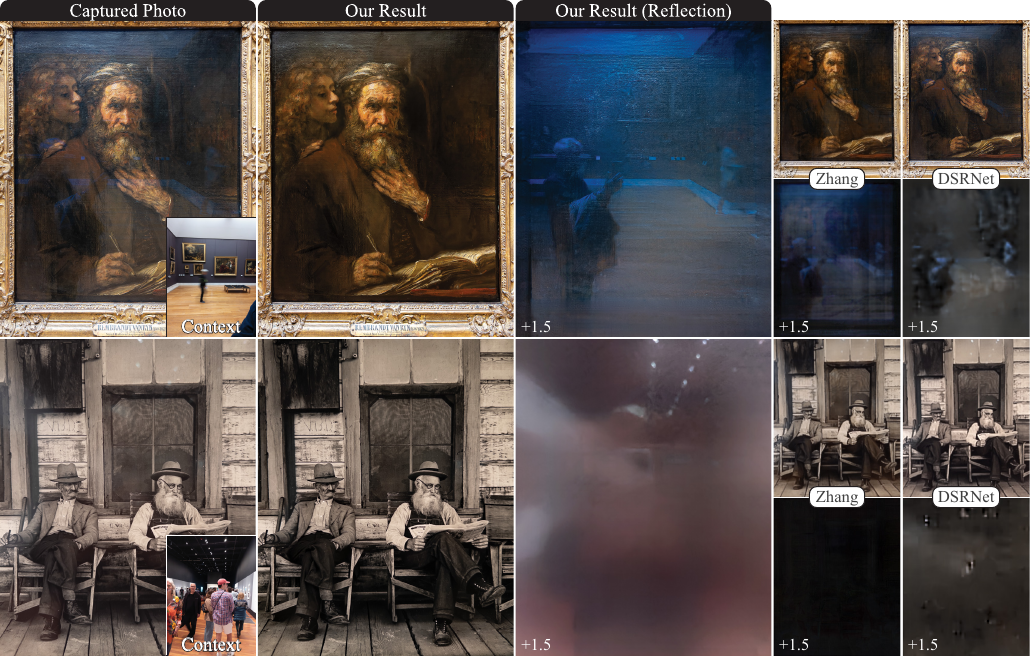}   
    \caption{Reflection recovery. Our model separates reflections that can be difficult to spot with the naked eye. See  \cref{sec:supp:subtle_reflections}.}
    \label{fig:supp:extra_result}
\end{figure*}%
We found that our model removes blue colored reflections consistently well, and we believe that this results from their commonness (outdoor illuminants are powerful and therefore frequently create reflections that appear blue when mixed with interior scenes). The yellow color of interior reflections on outdoor scenes also seems to help, as our model can separate textured objects like the painting on the wall in \cref{fig:gt_upcompare} from the tree texture. We also tried illuminating the indoor scene in \cref{fig:gt_upcompare} with a special studio light that is much more powerful than a typical interior light source. This created a warm indoor reflection that was analogous to the display case in \cref{fig:supp:results_basecompare} (right), where the reflection covers large parts of the transmission scene. Our model results are less consistent in this artificial situation. We believe this is because it is rare for interiors to be flooded with such strong lights, and so their reflections are uncommon in our training data. We find that interior reflections tend to appear in small regions because most artificial light sources are weak---only objects near the light will be bright enough to reflect over outdoor scenes. At nighttime, however, consumer illuminants easily reflect over dark cityscapes. We found that our model results are less consistent on such images. This can be improved by augmenting the dataset with source images $t = T(i)$ that were taken at nighttime. 
\begin{figure*}[t!]
  \centering
    \includegraphics[width=\linewidth]{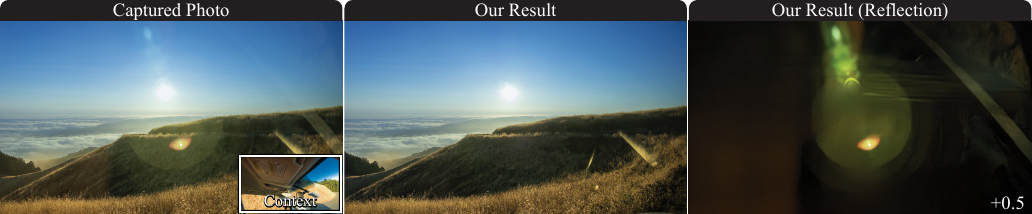}   
    \caption{Removing lens flares. Although our training data do not include images of lens flares, our model can sometimes remove them.}
    \label{fig:supp:flare_result}
\end{figure*}

\begin{figure*}[h!]
  \centering
    \includegraphics[width=\linewidth]{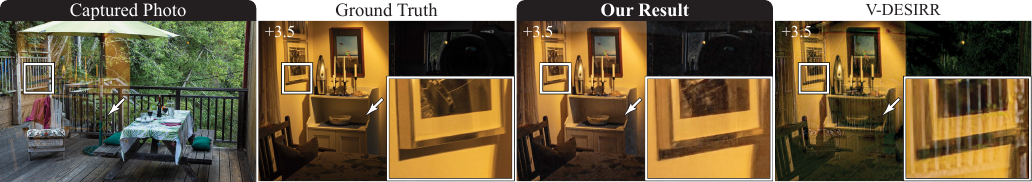}   
    \caption{Upsampler performance comparison. We upsample a ground truth reflection image from $256\times384$ to $ 2048\times3072$ using our method and V-DESIRR~\cite{prasad2021}. The latter creates strong artifacts due to resize operations that directly synthesize output pixels. Our model eliminates these artifacts while using $29\%$ fewer parameters.}
    \label{fig:supp:results_upcompare}
\end{figure*}

\textbf{\mbox{Reflections in the wild}}. In \cref{fig:supp:results_usecases}, \ref{fig:supp:results_usecases_more}, and \cref{fig:results_usecases} we show results on reflections that were captured in-the-wild, where it was not possible to capture ground truth: (left) shopping in the morning, (middle) looking into a building at midday in a city, and (right) photographing artwork in an outdoor mall (\cref{fig:supp:results_usecases}); while traveling and photographing artwork from a city street (\ref{fig:supp:results_usecases_more}). We compare to Zhang~\cite{zhang2018} and DSRNet~\cite{qiming2023} using their published $8$-bit models (bottom two rows). The $8$-bit models do not consistently remove in-the-wild reflections, with the exception that the method of Zhang et al.\ seems to have learned to remove blue colored content (left); see also \cref{fig:supp:extra_result} (top). We retrained these prior works on our photometrically accurate training data (marked ``retrained''), and they improve significantly. This suggests that the muted response of the $8$-bit models on in-the-wild reflections results from differences between their training data and real world reflections, and furthermore that the training process of prior works is insufficient: pre-training on photometrically inaccurate images, and fine-tuning on small datasets of ground truth reflections does not produce models that generalize as well. Looking closely, however, note that the retrained models do still introduce blur and colored artifacts. 

Our methods perform well across these diverse in-the-wild use cases. In \cref{fig:supp:results_usecases}, both our contextual and non-contextual models separate the strongly blue colored reflection, the complex cityscape reflection, and the artwork reflection (in the entrance to an indoor shopping mall). The contextual model improves each of these cases, excepting the transmission image for the shopping photo. We believe this also results from the saturated regions in the upper right of the contextual photo, where we have used the as-shot illuminant color in the EXIF to recover the pixels. 

The midday city photo represents a difficult reflection, and both of our models improve this image. The result of the contextual model is more correct: the interior mural is more accurately reconstructed, and the white sign that is adhered to the glass is left more intact. Saturated regions near the ceiling have color artifacts in both cases. 

Lastly, the artwork example (right) has two illuminants, a small warm colored artificial light, and a dominant outdoor illuminant. The white balance of the image is determined by the outdoor illuminant, which also illuminates the reflection scene, and this leads to a diversely colored reflection that both of our models are able to remove. The contextual model better preserves the texture of the artwork, and does not associate that texture with the reflection, whereas the non-contextual model does.

\textbf{\mbox{Subtle reflections}}.\label{sec:supp:subtle_reflections} In \cref{fig:supp:extra_result} we show results on images with subtle reflections, which are common in art museums where the glass and lighting are optimized to reduce reflections. Our models are able to remove these reflections, and recover $r$ even when it is invisible to the naked eye. In the painting (Rembrandt's ``St.\ Matthew and the Angel''), the recovered reflection has a strong blue color %
due to the special glass that is used in museums, and it correctly depicts the photographer and gallery. The colors in the recovered painting are also more correct. The specular reflection from the surface of the painting (top) is not removed, and we believe that this is the desired result. We have found that our models will sometimes remove sharp specular regions that are similar in color to the reflection from the glass, and this is visible in the upper quarter of the reflection image, where some specular texture from the surface of the painting is visible, and has been removed from the transmission.

In \cref{fig:supp:extra_result} (bottom) we remove reflections from Ansel Adams' ``Residents of Hornitos.'' Ceiling light sources are visible at the top of the photo, and on the left there is a general loss of contrast. The recovered transmission image has accurately uniform contrast, and the reflection reveals the hidden image of the photographer holding their cell phone. Our model is able to recover hidden reflections like this in part due to the bit depth of RAW images. The reflection is also blurry and differs in color.

\begin{figure*}[t]
	\centering
	\includegraphics[width=1\linewidth]{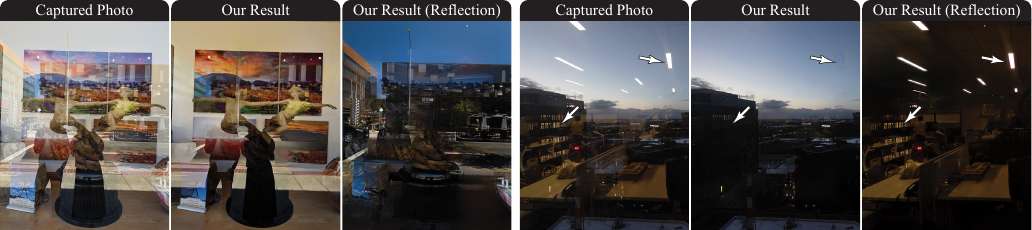}   
	\caption{Failure cases. Hard-to-interpret photos can cause reflection removal to be imperfect.}
	\label{fig:failures_base_model}
\end{figure*}

\begin{figure}[t]
	\centering
	\includegraphics{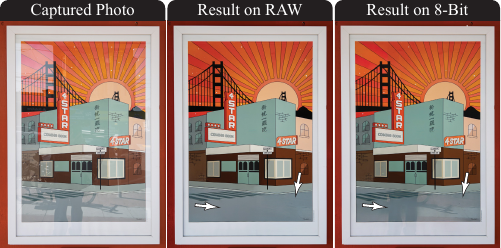}   
	\caption{Results when running a RAW-trained model on 8-bit inputs. Results on 8-bit photos can be obtained by pseudo-inverting gamma, but they are often imperfect because such photos have undergone a variety of non-linear finishing effects that are not present in RAW images (see ACR Steps \ref{acr:tone_mapping}-\ref{acr:gamma_compression}).}
	\label{fig:jpeg_compare}
\end{figure}

\begin{figure}[h!]
    \centering
    \includegraphics[width=0.925\linewidth]{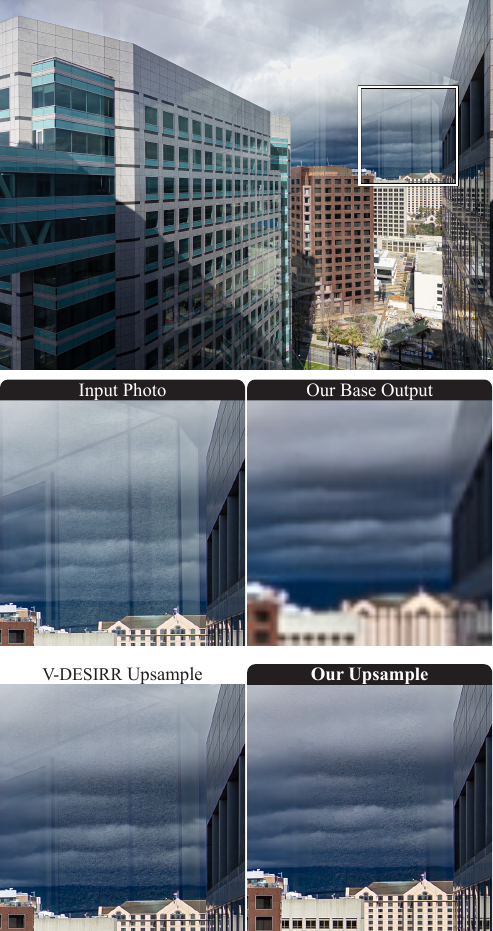}  
    \caption{Upsampling errors. V-DESIRR~\cite{prasad2021} re-introduces low-frequency reflections. Our method avoids this error, but copies high-frequency textures that were not visible at $256^2$.}
    \label{fig:supp:upsampler_failure}
    \vspace{-1em}
\end{figure}

\begin{figure*}[h!]
  \centering
    \includegraphics[width=\linewidth]{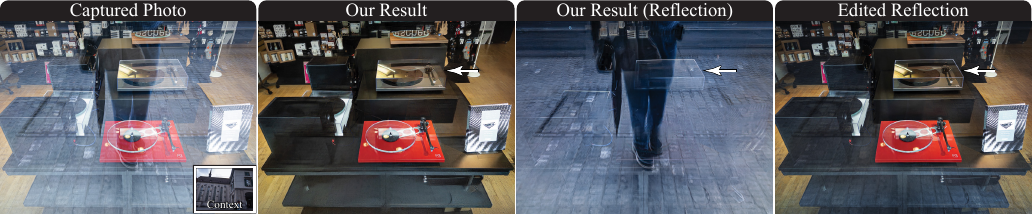}   
    \caption{Predicting the reflection enables aesthetic editing and error correction.}
    \label{fig:supp:editing_repair}
    \vspace{-1em}
\end{figure*}

Our ability to uncover subtle reflections is in part due to the extended bit depth of RAW images. The images in \cref{fig:main_result}, \cref{fig:results_upcompare}-\ref{fig:results_usecases}, \cref{fig:supp:results_basecompare}-\ref{fig:supp:editing_repair}  are $\ge$15-bit. ACR Step \ref{acr:subtract_black} is uint16. Together, the source datasets for the simulation (MIT5K~\cite{mit5k} and RAISE~\cite{raise}), are 12- and 14-bit (43\%, 57\%).%

\textbf{\mbox{Lens flares}}. Our model can also remove lens flares, which are reflections from the optical elements within the camera lens. An example is shown in \cref{fig:supp:flare_result}.  Most of the lens flare is removed, excepting one saturated region. The flare itself is also recovered in the reflection image even though the simulated reflections $r$ in the training dataset do not include lens flares. Since our method has some ability to generalize to flare removal, and it is difficult to create training data for flare removal, it could be helpful to pre-train flare removal methods first to remove photometrically accurate simulated reflections.

\textbf{Failures}. Failures typically occur when an image is too difficult for a person to visually interpret, as shown in \cref{fig:failures_base_model}. Light sources might also be missed, or a halo might be left behind, but in-painting methods can fill these holes. Reflections are usually removed best when glass covers the field-of-view, blocking the photographer from their subject, as this matches how the simulation is designed. As a result, local reflections on objects are typically not modified, which is desirable (\eg shiny cars, distant windows).%

\textbf{Results on 8-bit photos}.\label{sec:supp:8bit_results}
In this work, all results are presented on RAW photos because our models are trained for RAW. Nonetheless, our RAW-trained models have some ability to remove reflections in 8-bit photos if gamma is pseudo-inverted, as shown in \cref{fig:jpeg_compare}, but removal may be incomplete. This difference between RAW and 8-bit photos appears because the latter are produced by diverse camera pipelines and artistic adjustments. This diversity of finishing operations must be considered when training and testing models that have been trained on RAW-based simulation images. It is therefore not meaningful to test our models on datasets of 8-bit images.%


\subsection{Upsampler comparisons}\label{sec:supp:upsampler_comparisons}

In \cref{fig:supp:results_upcompare} we provide additional comparison to \mbox{V-DESIRR}~\cite{prasad2021} in which we upsample a ground truth reflection image (a transmission image is upsampled in \cref{fig:results_upcompare}). A magnified region is inset, and shows that our method preserves details of a framed photo through the upsampling process, whereas V-DESIRR re-introduces the reflection content (the bars of a fence; see captured photo, white box). In the rest of the image, V-DESIRR introduces strong color artifacts around sharp edges. This is a consistent issue that appears to result from the propagation and amplification of small errors that are made at low resolutions. Our model reduces error propagation by masking $t$ and $r$ out of the high-resolution mixture features at each level. We are able to predict an effective, high resolution mask with a lightweight and fast model by matching features, whereas V-DESIRR must use its model capacity to infer from the high-resolution mixture image what details to add into the low resolution clean images. This latter problem is more difficult to solve with limited model capacity, and it is difficult to avoid propagating errors.

In \cref{fig:supp:upsampler_failure} we show a typical failure mode of our model. The input image (top and center left) has a reflection of a textured exterior wall. Our base model correctly removes this reflection at $256^2$ pixels (center right).  Notice that, at low resolution, it is possible to represent the vertical edges of the exterior wall, but not the texture of the wall. We upsampled this result with both V-DESIRR and our model. %

V-DESIRR re-introduces the low frequency edges that were removed by the base model, and it copies the high frequency wall texture into the transmission as well. Our model does not re-introduce the low frequency edges, but the high frequency texture of the wall that is not present at low resolution is copied into the transmission image. This produces a texture artifact in the final result. Future work should reduce these kinds of errors while keeping the architecture lightweight. This might be done by reusing feature information from the base model, and across levels of the Gaussian pyramid as the upsampler is iteratively applied.

\subsection{Editing reflections}
\label{sec:supp:editing}

In \cref{fig:editing} and \cref{fig:supp:editing_repair} we show examples of reflection editing for aesthetic control and error correction (see also \cref{sec:editing}). For \cref{fig:supp:editing_repair}, a photographer was asked to finish the photo using the outputs of the reflection removal system. They chose to re-introduce the reflection for aesthetic purposes, and to correct errors. The reflections from the edges of the top record player cover were removed by our system (white arrows); the photographer pulled them back into their edited image. Editing was performed in Photoshop using the tone-mapped transmission and reflection images, and the ``Linear Dodge (Add)'' layer blend mode.

\section{Adobe Camera RAW, DNG SDK}\label{supp:sec:adobe_camera_raw}

Here we detail the code within the DNG SDK~\cite{acr} version 1.7.1 that is used in our simulation functions \fref{func:simulate_examples}, \fref{func:compute_exposure} and \fref{func:white_balance}. In this work, the necessary SDK functionality was transliterated into Python to interoperate with the geometric simulation (\cref{sec:supp:geometric_sim}) and mixture search (\cref{sec:supp:data_collection}). To simplify the exposition, we however describe the SDK code here in a functional manner, whereas the SDK is a class system. Consequently, some of the SDK functions use class member variables, and not all functions in this exposition of the code map one-to-one with functions of the same name in the SDK.

As discussed in \cref{sec:reflection_synthesis} and \cref{sec:supp:photometric_sim}, reflections are simulated in XYZ color space by using the color processing of the DNG SDK, which supports two paths to a white balanced, linear RGB image: the \ttbig{ForwardMatrix} or the \ttbig{ColorMatrix}. Only the latter path facilitates conversion to a device-independent color space (XYZ) before white balancing, as required for reflection simulation. We therefore implement in \fref{func:extract_xyz_images} the ACR color process in which the \ttbig{ColorMatrix} is used (cf.~\mbox{\fref{func:cam_to_rgb}}). Note that both paths account for the as-shot illuminant because the \ttbig{ColorMatrix} is interpolated according to the as-shot illuminant (see \fref{func:calc_whitexy} and \fref{func:find_xyz_to_cam}). 

All supporting DNG SDK functions are listed below.

\begin{algorithm}[t]
\algofontsize{}
\caption{Adobe DNG SDK, convert RAW images to XYZ.}
\footnotesize
\textbf{Input:} A RAW image\\
\mbox{\textbf{Output:} A linear image in XYZ color}%
\begin{algorithmic}[1]
  \footnotesize
    \STATE Extract the ACR Stage-3 image \tt{I} with \tt{dng\_validate} option \tt{-3}. 
    \\\hfill\COMMENT{ACR Step~\ref{acr:demosaic}}
    \STATE Subtract the Stage-3 black level from \tt{I}. \hfill\COMMENT{SDK \fref{func:stage3_black_level}}
    \STATE Divide \tt{I} by the maximum pixel value.
    \STATE Compute \tt{WhiteXY}. \hfill\COMMENT{SDK \fref{func:calc_whitexy}}
    \STATE Compute the transform \tt{XYZ\_to\_CAM} from \tt{WhiteXY}. \hfill\COMMENT{SDK \fref{func:find_xyz_to_cam}}    
    \STATE Recover saturated highlights in \tt{I}. \hfill\COMMENT{SDK \fref{func:highlight_recovery}}
    \STATE Transform \tt{I} to XYZ using \tt{inv(XYZ\_to\_CAM)}. 
    \\\hfill\COMMENT{see also SDK \fref{func:cam_to_rgb}}
    \RETURN the linear XYZ image \tt{I}.
\end{algorithmic}
\label{func:extract_xyz_images}
\end{algorithm}

\begin{algorithm}[p]
\algofontsize{}
\caption{Adobe DNG SDK, \ttbig{CAM\_to\_RGB}}
\footnotesize
\textbf{Input:} \tt{WhiteXY} \\
\textbf{Output:} Transform to linear RGB\\\\[-0.5em]
This function is included here as reference to the entire computation of the color transform to linear RGB. Note, the DNG SDK implements the DNG Spec.~\cite{acr}. It uses the \tt{ForwardMatrix} when it is available, and otherwise uses the \tt{CameraMatrix}. In this work we use only the \tt{CameraMatrix}, since this supports white balancing after conversion to XYZ. \\[-0.5em]
    \begin{algorithmic}[1]
    \STATE See \tt{dng\_color\_spec.cpp:573-609}.
\end{algorithmic}
\label{func:cam_to_rgb}
\end{algorithm}
\begin{algorithm}[p]
\algofontsize{}
\caption{Adobe DNG SDK, get \ttbig{Stage3BlackLevel}}
\footnotesize
\textbf{Input:} A DNG file \\
\textbf{Output:} The black level \\\\[-0.5em]
{\em The Stage-3 black level is not stored in the DNG EXIF header.} It is a global scalar offset that remains after spatially varying black levels have been removed by parsing and applying the black level tags. By default, the DNG SDK Stage-3 image has a black level of zero, and negative noise values have been clipped to zero. In this work, we adopt that convention and clip the negative noise for simplicity. Clipping can however be disabled, and the non-zero Stage-3 black level can be recovered with a minor modification to the \tt{dng\_validate} binary, described below. The black level can then be read from the printed output of \tt{dng\_validate} when the stage-3 image is extracted with the \tt{-3} option.\\[-0.5em]
\begin{algorithmic}[1]
	\STATE \tt{return true} for \tt{SupportsPreservedBlackLevels} \\\hfill\COMMENT{\tt{dng\_mosaic\_info.cpp:2014}}
	\STATE \tt{return true} for \tt{SupportsPreservedBlackLevels} \\\hfill\COMMENT{\tt{dng\_negative.cpp:5814}}
    \STATE \tt{printf((uint16) negative->Stage3BlackLevel())} \\\hfill\COMMENT{\tt{dng\_validate.cpp:293}}
\end{algorithmic}
\label{func:stage3_black_level}
\end{algorithm}
\begin{algorithm}[p]
\algofontsize{}
\caption{Adobe DNG SDK, recover saturated highlights}
\footnotesize
\textbf{Input:} An image in camera color space \\
\textbf{Output:} Image with clipped higlights repaired
\begin{algorithmic}[1]
    \STATE Compute \tt{CameraWhite} \\\hfill\COMMENT{\tt{dng\_color\_spec.cpp:548-568}}%
    \RETURN $\forall c,\min(c,~$\tt{CameraWhite}$)$ \\\hfill\COMMENT{\tt{dng\_render.cpp:1785} $\rightarrow$ \tt{dng\_reference.cpp:1389}}
\end{algorithmic}
\label{func:highlight_recovery}
\end{algorithm}

\begin{algorithm}[p]
\algofontsize{}
\caption{Adobe DNG SDK, \ttbig{FindXYZtoCamera}}
\footnotesize
\textbf{Input:}  White point XY \\
\textbf{Output:} Matrix \tt{XYZ\_to\_CAM}
\begin{algorithmic}[1]
	\STATE See \tt{dng\_color\_spec.cpp:541} \\\hfill\COMMENT{In practice, calls \tt{FindXYZtoCamera\_SingleOrDual}.}
\end{algorithmic}
\label{func:find_xyz_to_cam}
\end{algorithm}

\begin{algorithm}[p]
\algofontsize{}
\caption{Adobe DNG SDK, \ttbig{NeutralToXY} (projected, cf. SDK \fref{func:neutral_to_xy})}
\footnotesize
\textbf{Input:} An AWB white point in camera color space. \\
\textbf{Output:} \tt{XYZ\_to\_CAM\_awb} and \tt{WhiteXY\_awb} 
\begin{algorithmic}[1]
	\FORALL[plausible kelvin values]{kelvins $K$}
		\STATE Compute the XY coordinate of $K$. \hfill\COMMENT{SDK \fref{func:get_xy_coord}}
		\STATE Compute the transform \tt{XYZ\_to\_CAM} from XY. \hfill\COMMENT{SDK \fref{func:find_xyz_to_cam}}
		\STATE Compute the XYZ coordinate of XY. \hfill\COMMENT{SDK \fref{func:xy_to_xyz}}
		\STATE Project the XYZ coordinate into camera color using \tt{XYZ\_to\_CAM}.
		\IF{the projected XYZ is closer to the AWB point than previous values}
			\STATE Save XY
		\ENDIF
	\ENDFOR
	\RETURN the saved XY value and its associated \tt{XYZ\_to\_CAM} matrix.
\end{algorithmic}
\label{func:xyz_to_cam_awb}
\end{algorithm}

\begin{algorithm}[p]
\algofontsize{}
\caption{Adobe DNG SDK, computing \ttbig{WhiteXY} and \ttbig{CameraWhite}}
\footnotesize
\textbf{Input:}  DNG \tt{AsShotXY} XOR \tt{AsShotNeutral} \{All DNGs have one xor the other.\}\\
\textbf{Output:}  White point in XY
\begin{algorithmic}[1]
	\STATE Compute \tt{WhiteXY}. \hfill\COMMENT{\tt{dng\_render.cpp:892,899}}%
\end{algorithmic}
\label{func:calc_whitexy}
\end{algorithm}

\begin{algorithm}[p]
\algofontsize{}
\caption{Adobe DNG SDK, \ttbig{NeutralToXY}}
\footnotesize
\textbf{Input:}  DNG \tt{AsShotNeutral} value \\
\textbf{Output:} An XY coordinate
\begin{algorithmic}[1] 
	\STATE See \tt{dng\_color\_spec.cpp:659}
\end{algorithmic}
\label{func:neutral_to_xy}
\end{algorithm}

\begin{algorithm}[p]
\algofontsize{}
\caption{Adobe DNG SDK, compute \ttbig{XYZ\_to\_sRGB}}
\footnotesize
\textbf{Input:}  An XY white point, \tt{XYPoint}. \\
\textbf{Output:}  Matrix \tt{XYZ\_to\_sRGB}.
\begin{algorithmic}[1]
    \STATE Get the transform \tt{XYZ\_D50\_to\_sRGB}. \hfill\COMMENT{SDK \fref{func:xyz_d50_to_srgb}}
    \STATE Get the D50 XY coordinate \tt{XY\_D50}. \hfill\COMMENT{SDK \fref{func:d50_xy_coord}}
	\STATE Compute matrix \tt{XYZ\_to\_XYZ\_D50} using \tt{XY\_D50} and \tt{XYPoint}. \\\hfill\COMMENT{SDK \fref{func:map_white_matrix}}
    \RETURN \tt{XYZ\_D50\_to\_sRGB} $\boldsymbol{\cdot}$ \tt{XYZ\_to\_XYZ\_D50}
\end{algorithmic}
\label{func:XYZ_to_linear_sRGB}
\end{algorithm}

\begin{algorithm}[p]
\algofontsize{}
\caption{Adobe DNG SDK, get \ttbig{XYZ\_D50\_to\_sRGB}}
\footnotesize
\textbf{Input:}  None \\
\mbox{\textbf{Output:} The transform from XYZ D50 to linear sRGB.}
\begin{algorithmic}[1]
	\STATE See \tt{dng\_color\_space.cpp:254}, which specifies the inverse matrix.
\end{algorithmic}
\label{func:xyz_d50_to_srgb}
\end{algorithm}

\begin{algorithm}[p]
\algofontsize{}
\caption{Adobe DNG SDK, \ttbig{MapWhiteMatrix}}
\footnotesize
\textbf{Input:}  Two white points, \tt{w1} to \tt{w2}. \\
\mbox{\textbf{Output:} Bradford adaptation matrix.}
\begin{algorithmic}[1]
	\STATE See \tt{dng\_color\_spec.cpp:22}.
\end{algorithmic}
\label{func:map_white_matrix}
\end{algorithm}

\begin{algorithm}[p]
\algofontsize{}
\caption{Adobe DNG SDK, \ttbig{D50\_xy\_coord}}
\footnotesize
\textbf{Input:}  None \\
\textbf{Output:} XY coordinate of the D50 illuminant
\begin{algorithmic}[1]
	\STATE See \tt{dng\_xy\_coord.h:145}.
\end{algorithmic}
\label{func:d50_xy_coord}
\end{algorithm}

\begin{algorithm}[p]
\algofontsize{}
\caption{Adobe DNG SDK, \ttbig{Get\_xy\_coord}.}
\footnotesize
\textbf{Input:} A scalar temperature value, $K$. \\
\textbf{Output:} An XY coordinate.
\begin{algorithmic}[1]
	\STATE See \tt{dng\_temperature.cpp:173}.
\end{algorithmic}
\label{func:get_xy_coord}
\end{algorithm}

\begin{algorithm}[p]
\algofontsize{}
\caption{Adobe DNG SDK, \ttbig{XYtoXYZ}.}
\footnotesize
\textbf{Input:} An XY vaue. \\
\textbf{Output:} An XYZ value
\begin{algorithmic}[1]
	\STATE See \tt{dng\_xy\_coord.cpp:47}.
\end{algorithmic}
\label{func:xy_to_xyz}
\end{algorithm}

\begin{algorithm}[p]
\algofontsize{}
\caption{Adobe DNG SDK, \ttbig{sRGB\_to\_linear\_sRGB}.}
\footnotesize
\textbf{Input:} A gamma compressed sRGB value \\
\textbf{Output:} A linear sRGB value
\begin{algorithmic}[1]
	\STATE See \tt{dng\_color\_space.cpp:34}.
\end{algorithmic}
\label{func:srgb_to_linear_srgb}
\end{algorithm}
\end{appendices}

\end{document}